\documentclass[letterpaper]{article} 
\usepackage{aaai2026}  

\usepackage{times}  
\usepackage{helvet}  
\usepackage{courier}  
\usepackage[hyphens]{url}  
\usepackage{graphicx} 
\usepackage{natbib}  
\usepackage{caption} 
\usepackage{algorithm}
\usepackage{algorithmic}

\usepackage{newfloat}
\usepackage{listings}
\DeclareCaptionStyle{ruled}{labelfont=normalfont,labelsep=colon,strut=off} 
\floatstyle{ruled}
\newfloat{listing}{tb}{lst}{}
\floatname{listing}{Listing}
\usepackage{subcaption}
\usepackage{multirow}

\usepackage{amssymb}
\usepackage{amsmath}
\usepackage{pifont}

\usepackage{xcolor}

\title{Achieving Fairness Without Harm via Selective Demographic Experts} 

\author {
   Xuwei Tan\textsuperscript{\rm 1},
   Yuanlong Wang\textsuperscript{\rm 1,2},
   Thai-Hoang Pham\textsuperscript{\rm 1,2},
   Ping Zhang\textsuperscript{\rm 1,2},
    Xueru Zhang\textsuperscript{\rm 1}
}
\affiliations {
    \textsuperscript{\rm 1}Department of Computer Science and Engineering, The Ohio State University, USA\\
    \textsuperscript{\rm 2}Department of Biomedical Informatics, The Ohio State University, USA\\
    \{tan.1206, wang.16050, pham.375, zhang.10631, zhang.12807\}@osu.edu
}

\begin{document}

\maketitle

\begin{abstract}

As machine learning systems become increasingly integrated into human-centered domains such as healthcare, ensuring fairness while maintaining high predictive performance is critical. Existing bias mitigation techniques often impose a trade-off between fairness and accuracy, inadvertently degrading performance for certain demographic groups. In high-stakes domains like clinical diagnosis, such trade-offs are ethically and practically unacceptable. In this study, we propose a fairness-without-harm approach by learning distinct representations for different demographic groups and selectively applying demographic experts consisting of group-specific representations and personalized classifiers through a no-harm constrained selection. We evaluate our approach on three real-world medical datasets---covering eye disease, skin cancer, and X-ray diagnosis---as well as two face datasets. Extensive empirical results demonstrate the effectiveness of our approach in achieving fairness without harm.

\end{abstract}

\begin{links}
    \link{Code}{https://github.com/osu-srml/FairSDE}
\end{links}

\section{Introduction}
\begin{figure}[t]
    \centering
        \includegraphics[width=0.9\linewidth]{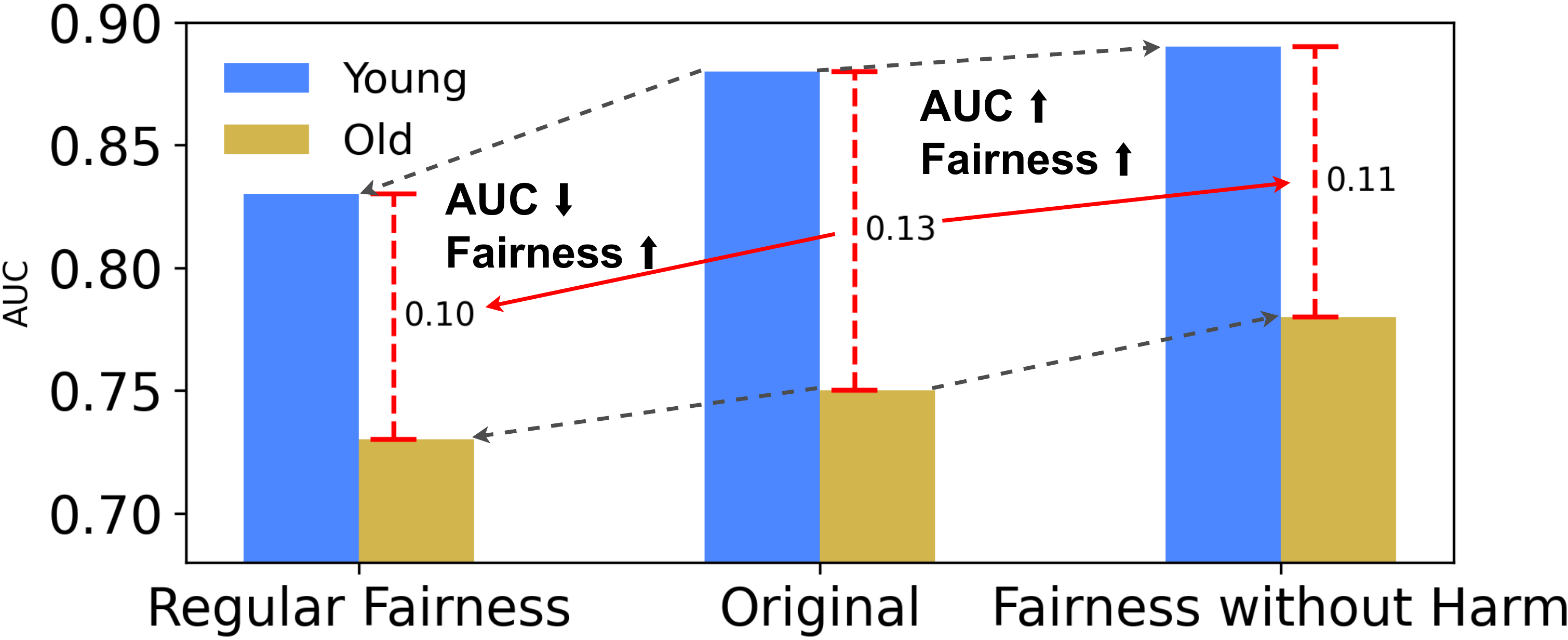}
    \caption{AUC of ResNet-18 on \texttt{Ham10000} for malignant prediction using age as the sensitive group. The 13\% gap in AUC between age groups indicates bias.  Regular approaches that enforce group constraints (e.g., left \textit{Young} and \textit{Old}) reduce this gap but degrade performance for both groups. In contrast, our method improves fairness without harming either group's performance, as seen in right \textit{Young} and \textit{Old}. } 
    \label{fig:unfair_demo}
\end{figure}

As machine learning (ML) is increasingly used in high-stakes domains like healthcare, finance, and criminal justice, concerns about algorithmic bias and its impact on marginalized groups have grown. In medical domain, fairness is particularly critical: diagnostic tools that underperform on certain age, gender, or racial groups can lead to misdiagnoses, delayed treatments, and ultimately, harm to patients. For instance, a model developed to diagnose skin diseases \cite{maron2019systematic} was shown to exhibit age bias, with up to a 13\% gap in area under the curve (AUC) between younger and older patients (Figure~\ref{fig:unfair_demo}). Such performance disparities highlight the urgent need for effective bias mitigation strategies to promote equitable outcomes across diverse social groups.

To quantify model unfairness, existing studies have proposed many notions of \textbf{group fairness}, which require statistical measures (e.g., accuracy, true/false positive rate) to be equal across different groups. Commonly used notions include \textit{demographic parity} \cite{dwork2012fairness}, \textit{equal opportunity} and \textit{equalized odds} \cite{hardt2016equality}. Based on them, various approaches have been developed and can be broadly categorized into three types: (i) \textbf{\textit{pre-processing}}, by modifying the
original dataset such as removing certain features or reweighing samples \cite{kamiran2012data,zemel2013learning,gordaliza2019obtaining}; (ii) \textbf{\textit{in-processing}}, by modifying the learning algorithms such as imposing fairness constraints or changing objective functions \cite{zafar2019fairness,zafar2017fairness,agarwal2018reductions}; (iii) \textbf{\textit{post-processing}}, by adjusting model outputs based on sensitive attributes \cite{hardt2016equality,khalili2021fair}. However, these methods often enhance fairness at the cost of reduced accuracy. Notably, performance deterioration may happen to all groups, including those disadvantaged (see Figure~\ref{fig:unfair_demo} for an illustration). 

In high-stakes domains like healthcare, sacrificing any group's model performance for fairness is unacceptable, as it conflicts with the ethical principles of \textit{beneficence} (doing good) and \textit{non-maleficence} (avoiding harm) \cite{beauchamp1994principles}. A more desirable goal is to improve outcomes for disadvantaged groups without compromising performance for others, as illustrated in Figure~\ref{fig:unfair_demo}.

This paper aims to achieve fairness in ML without compromising the model performance of any group. The most closely related works are \cite{martinez2019fairness,ustun2019fairness,yin2024fair,cai2025demographic}. Specifically, \citet{martinez2019fairness} focused on finding a \textit{Pareto-optimal} fair model that minimizes the performance gap among different groups without \textit{unnecessary} harm (i.e., minimal accuracy reduction for any group); this differs from this paper where we aim to avoid performance degradation for any group. To prevent harm,
\citet{ustun2019fairness} suggested utilizing individuals' demographic information (i.e., sensitive attributes) to train a set of \textit{decoupled} models. Rather than equalizing outcomes across different groups, their goal is to ensure that each group achieves the best performance with its assigned decoupled model, compared to the pooled model (trained on all groups) and the decoupled models of other groups. More recently, \cite{cai2025demographic} considered a similar fairness notion, but introduced a demographic-agnostic method for learning decoupled models. Another study \cite{yin2024fair} proposed a post-processing method that utilizes \textit{abstention} to achieve group fairness without harming accuracy, i.e., adjusting the output of a pre-trained ML model that selectively abstains from making predictions on certain samples, deferring decisions to humans. However, this approach requires human involvement and has limited scalability. 

Moreover, methods introduced in \cite{martinez2019fairness,ustun2019fairness,yin2024fair,cai2025demographic} were primarily evaluated on simple tabular data, such as \texttt{Adult} \cite{asuncion2007uci}, \texttt{COMPAS} \cite{bellamy2018ai}, and \texttt{Law} \cite{bellamy2018ai}, while their performance on high-dimensional image data remains less explored. As we will demonstrate in this paper, high-dimensional features, such as medical images, are often not easily separable by sensitive attributes in the latent space, making it highly challenging to achieve fairness without harm using decoupled or abstained classifiers.

In this paper, we propose methods that \textbf{improve model fairness without reducing performance for any group}, relative to an unconstrained pooled empirical risk minimization (ERM) baseline trained on the full population. Our approach is well-suited for \textbf{high-dimensional} feature spaces. Inspired by \citet{ustun2019fairness}, we adopt personalized classifiers for each demographic group to achieve fairness without harm. The use of demographic information is motivated by real-world medical practice, where attributes such as age, sex, or ethnicity routinely inform diagnosis and treatment for specific subpopulations. A key challenge is that naively personalized classifiers may underperform pooled ERM when group distributions are similar, as the pooled model benefits from more data to learn shared patterns. To address this, we learn \textit{demographic experts}—group-specific representations paired with personalized classifiers—that better capture group distributional differences. Building on experts, we develop a post-processing method that dynamically selects between expert models and the pooled model via a \textit{fairness without harm} guided combinatorial optimization. Our main contributions are:

\begin{itemize}
    \item  We explore the feasibility of achieving fairness without harm on complex image datasets, addressing a critical gap in current research, which has primarily focused on simpler tabular datasets.
    \item We propose \textbf{FairSDE}, an extension of decoupled classifiers to  \textbf{Fair} \textbf{S}elective \textbf{D}emographic \textbf{E}xperts, which decouples both representations and classifiers to enable group-specific models to effectively capture group-specific patterns. Then, we selectively adjust predictions by dynamically choosing between the expert and pooled models via a combinatorial optimization framework.
    \item We conduct extensive experiments on real-world medical imaging data, demonstrating that FairSDE consistently achieves fairness without harm. This highlights its potential for practical deployment in performance-critical applications.  In contrast, existing methods often achieve fairness by sacrificing performance for certain groups.
\end{itemize}

\section{Related Work}
\noindent \textbf{Fairness notions.} Various notions have been proposed to measure the algorithmic unfairness, which can be broadly classified into the following. \textit{Unawareness} prohibits the use of sensitive attributes in the model training and decision-making process. \textit{Parity-based fairness} requires certain statistical measures to be equalized across different groups, including Demographic Parity \cite{dwork2012fairness}, Equal Opportunity \cite{hardt2016equality}, Equalized Odds  \cite{hardt2016equality,pham2023fairness}, Predictive Parity \cite{chouldechova2017fair}, Accuracy Parity \cite{khalili2023loss,zhang2019group}, etc. \textit{Preference-based fairness}, inspired by fair division and envy-freeness in economics, ensures that each group prefers its own treatment over others, regardless of inter-group disparities \cite{zafar2017parity,ustun2019fairness}. \textit{Counterfactual fairness} holds when an individual’s outcome remains unchanged in a hypothetical world where their sensitive attribute is different \cite{kusner2017counterfactual,zuo2024lookahead,zuo2023counterfactually}.
Finally, \textit{individual fairness} treats similar individuals similarly for individual level fairness \cite{dwork2012fairness}.

\noindent \textbf{Approaches to mitigating unfairness.}
Many fairness algorithms have been proposed to address bias, which can be categorized into pre-processing, in-processing, and post-processing. Pre-processing methods manipulate datasets to mitigate bias in the data, such as reweighing \cite{kamiran2012data}, resampling \cite{kamiran2012data}, and data preprocessing \cite{celis2020data}. In-processing mitigation methods refer to regularizing the objective to guide the model in learning a fair classification. For example, adversarial training \cite{zhang2018mitigating, han2021diverse, han2021decoupling} learn a sensitive discriminator and reverse gradient to learn the group-invariant classification. \citet{shen2021contrastive, park2022fair} utilize contrastive learning to align the data from the same group and \citet{creager2019flexibly, park2021learning, lee2021learning} disentangle features to de-bias the model. Post-processing methods alter the prediction results to improve the fairness of the model. In this kind of method, \citet{hardt2016equality} creates separate thresholds for each sensitive group and alters the results to satisfy the specified fairness criteria, which results in lower accuracy in many cases. \citet{yin2024fair} train surrogate models based on the results of the baseline model to make further predictions.

\section{Problem Statement}

Let $\mathcal{S}$ be a dataset of $n$ individuals, where each individual $i$ is represented as $(x_i, y_i, a_i)$ drawn from the joint distribution $P(X, Y, A)$.  Here, $x_i = [x_{i,1}, \dots, x_{i,d}] \in \mathbb{R}^d$ is a $d$-dimensional feature vector, $y_i \in \mathcal{Y}$ is the label, and $a_i \in \mathcal{A}$ is a sensitive attribute (e.g., gender, race, age). We aim to learn a \textit{representation function} $f: \mathbb{R}^d \rightarrow \mathbb{R}^m$ that maps inputs to representations $z_i = f(x_i)$, and a set of group-specific \textit{classifiers} $h_a: \mathbb{R}^m \rightarrow \mathcal{Y}$ such that $\hat{y}_i = h_a(z_i)$ for individuals in group $a \in \mathcal{A}$. We use capital letters for random variables and lowercase for their realizations.

Let $R:\mathcal{Y}\times \mathcal{Y}\to \mathbb{R}_+$ be the risk/loss function measuring the discrepancy between prediction and ground truth, and $n_a$ be the number of samples in group $a\in\mathcal{A}$. Our goal is to learn a representation function $f$ and group-specific classifiers $\{h_a\}_{a\in\mathcal{A}}$ that minimize the risk of the entire population while ensuring fairness without harm: 
\begin{equation}
\begin{aligned}
    \text{minimize}~~~~ & \mathbb{E}_{A} \left[ \mathbb{E}_{X, Y | A} \left[ R \left( h_A \left( f(X) \right), Y \right) \right] \right] \\
    \text{s.t.}~~~~ & \text{No-harm constraint} \\
    ~~~~& \text{Fairness constraint}
\end{aligned}
\end{equation}

\noindent \textbf{No-harm constraint.} Denote $h_{\text{erm}}:\mathbb{R}^d\to \mathcal{Y}$ as the \begin{equation}
h_{\text{erm}} = \arg\min_{h \in \mathcal{H}} \sum\nolimits_{i=1}^n R(h(x_i), y_i)
\end{equation}
{where $\mathcal{H}$ is the hypothesis class.} It defines the baseline for the ``no harm" criterion: predictions using the representation function $f$ and  group-specific classifiers $\{h_a\}_{a\in\mathcal{A}}$ cause no harm if every group’s loss is no greater than that under $h_{\text{erm}}$, i.e., $\forall a\in \mathcal{A}$, the following should hold:
\begin{equation}
\resizebox{1\hsize}{!}{$\displaystyle    \mathbb{E}_{X, Y | A = a} \left[ R\left( h_a\left( f(X) \right), Y \right) \right] \leq \mathbb{E}_{X, Y | A = a} \left[ R\left( h_{\text{erm}}(X), Y \right) \right]$}
\end{equation}

\noindent \textbf{Fairness constraint.} We adopt two widely used metrics:
\begin{itemize}

\item \textit{Overall accuracy parity} \cite{berk2021fairness} requires similar overall accuracy across all groups.:
\begin{equation}
 \mathbb{P}[\hat{Y} = Y \mid A = a] = \mathbb{P}[\hat{Y} = Y \mid A = a'], \quad \forall a, a' \in \mathcal{A} 
\end{equation}

\item \textit{Max-min fairness} \cite{lahoti2020fairness} maximizes the worst-group performance to reduce disparities across groups. It is formally defined as:

\begin{equation}
    \max_{a \in \mathcal{A}} \mathbb{E}_{X, Y | A = a} \left[ R\left( h_a\left( f(X) \right), Y \right) \right] 
\end{equation}

\end{itemize}

\section{Methodology}

\begin{figure*}

    \centering
    \includegraphics[width=0.85\linewidth]{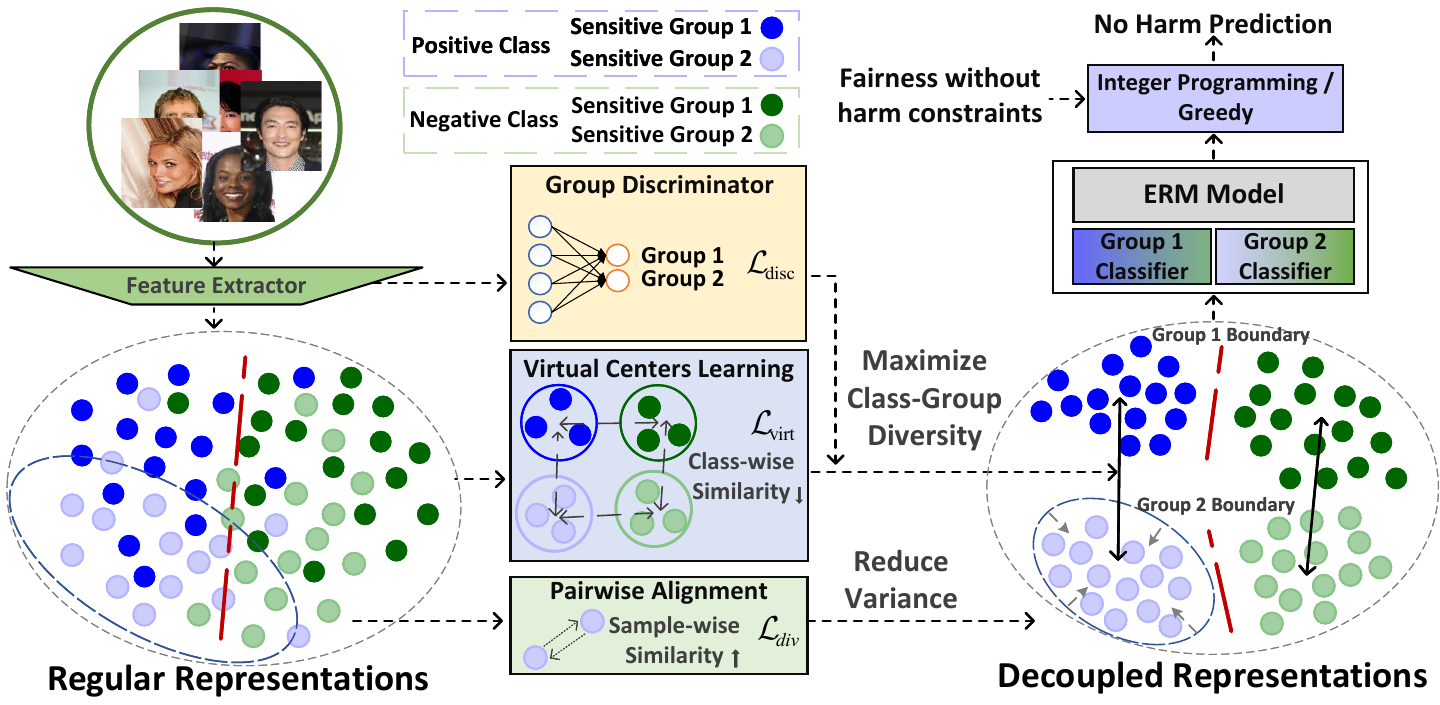}
    \caption{The illustration of FairSDE.  Decoupled Representations aims to create distinct representations for diverse sensitive groups by ensuring inner-group similarity and inter-group and inter-class diversity. In particular, we estimate the mean of the representation distribution using virtual centers and mutually align them to the samples from a specific class and group to ensure the representations of different groups are well-separated in the latent space. Additionally, by minimizing the pairwise distances within each group, we achieve a more compact group representation, leading to reduced variance. }
    \label{fig:algorithm}
\end{figure*}

Next, we present our approach to achieving fairness without harm. We first describe how to decouple representations to learn \textit{demographic experts}, followed by a dynamic expert selection method that adjusts predictions to enforce fairness. An overview is shown in Figure~\ref{fig:algorithm}.

\subsection{Decoupling Representations for Experts}
To advance group fairness while preserving task-specific discriminability, we propose a structured framework that learns demographic experts—specialized representations and classifiers for each subgroup defined by the sensitive attribute $A$ and the class $Y$. Unlike adversarial methods, which aim to eliminate the dependence between the sensitive attribute and certain variables (e.g., achieving accuracy parity by learning a predictor independent of the sensitive attribute), our approach explicitly models group-class distributions through expert representations. This ensures separability in latent space while maintaining intra-group cohesion. We optimize this objective at three levels: group-wise, class-wise, and sample-wise, achieving a structured and hierarchical disentanglement of representations.

Specifically, we explicitly enforce a \textit{group-wise} dependence between the representation $f(X)$ and the sensitive attribute $A$, i.e., by maximizing $P(A|f(X))$. This is achieved by introducing a discriminator $D: \mathbb{R}^m \to [0,1]^{|\mathcal{A}|}$ that predicts $A$ given the representation $f(X)$. We minimize $\mathcal{L}_{\text{disc}}$ to \textbf{link the representation $f(X)$ with its sensitive attribute $A$}, as defined below.
\begin{equation}
\mathcal{L}_{\text{disc}} = - \sum_{i=1}^{n} \sum_{a \in \mathcal{A}} \mathbb{I}(a_i = a) \log D(f(x_i))_a
\end{equation}

With the enforced group-wise dependence between representations and sensitive attributes, we then learn separable demographic representations that effectively diversify the representations across different groups. However, directly computing the representation distributions in the latent space is challenging. To address this, we introduce \textit{virtual centers} $V_{a,y}$ at class-wise level as proxies for the representation distributions. For each class $y$ and each sensitive group $a$, a virtual center is learned to represent the mean of the representation distribution. The similarity between each $V_{a,y}$ and the representation $f(x_i)$ of a sample from its respective class and sensitive group is measured by:
\begin{equation}
    d\left(V_{a,y}, f(x_i)\right) = \frac{V_{a,y} \cdot f(x_i)}{\|V_{a,y}\|  \|f(x_i)\|}
\end{equation}

By mutually aligning these similarities, we ensure that the virtual centers accurately reflect the central tendencies of their respective representation distributions and draw the related samples closer to these centers in the latent space. This alignment is achieved by maximizing bi-directional similarity, which is equal to minimizing $\mathcal{L}_{\text{virt}}$:
\begin{equation} 
\resizebox{0.9\hsize}{!}{$\displaystyle \mathcal{L}_{\text{virt}} = - \sum_{i=1}^n \sum_{a \in \mathcal{A}} \log \left( \frac{\exp(d(V_{a,y}, f(x_i)))}{\sum_{y' \in \mathcal{Y}} \exp(d(V_{a,y'}, f(x_i)))} \right)$}
\end{equation}
where  $\mathcal{Y}$ denotes all possible classes, respectively. By incorporating these virtual centers and aligning the similarities, we achieve a dual objective: (i) the virtual centers accurately represent the distribution centers, and (ii) the samples are naturally drawn toward these centers. As we associate representations with sensitive attributes, we can effectively learn the distribution for each class-group pair. 

To ensure distinct representations across different groups, we aim for these distributions to exhibit distinct means. Beyond mean divergence, variations in variance also influence the separability of the distributions. Specifically, if the distributions have low variance and distinct means, they will be more easily distinguishable in latent space. Formally, given the estimated center $V_{a,y}$ for each group and class, our goal is to maximize the distance between these centers while minimizing the number of samples located near the boundaries of the distributions. For each sample $x_i$, we penalize its similarity to centers from other classes or groups, thereby enhancing the diversity among the estimated centers. Furthermore, we introduce pairwise representation alignment to reduce the variance of the representation distributions. Specifically, we randomly select a sample $x'_i$ from the same class $y_i$ and group $a_i$, and a sample $x_i^{-}$ from a different class or group, optimizing both pairwise representation alignment and center diversity simultaneously:
\begin{equation}
    \resizebox{1\hsize}{!}{$\displaystyle \mathcal{L}_{\text{div}} = - \sum_{i=1}^n \log  \frac{\exp\left(f(x_i) \cdot f(x_i')\right) + \exp\left(d\left(V_{a_i,y_i}, f(x_i)\right)\right)}{\exp\left(f(x_i) \cdot f(x_i^{-})\right) + \displaystyle\sum_{
    a \neq a_i, y \neq y_i } \exp\left(d\left(V_{a,y}, f(x_i)\right)\right)}$} 
\end{equation}

Through random pairwise alignment and similarity penalization, we can better \textbf{diversify and compact the representation distribution}, particularly by pushing samples with high variance toward the center, which effectively reduces the variance of the representation distribution. This process results in separable representations that are more distinct and well-defined for each group and class.

\subsection{{Experts Dynamic Selection}}

With separable representations, we then train decoupled group-specific classifiers $\{h_a\}_{a \in \mathcal{A}}$ using samples from their respective groups to build the demographic experts. Each classifier is trained using the standard cross-entropy loss. For each group, we form the expert by coupling the group-specific representation with the corresponding classifier $h_a$. However, using each expert independently can be problematic, particularly for groups with limited data, as these classifiers may overfit or lack sufficient information to generalize effectively, which could result in worse performance or amplified bias. To address these challenges and ensure fairness without harm, we introduce a heuristic \textit{No-Harm Selection} to dynamically select either the experts or the pooled model. Following \citet{dutt2023fairtune}, we assume that the validation and test sets share a similar distribution. We evaluate our models on the validation set and adopt two strategies to satisfy max-min fairness or overall accuracy parity fairness.

\noindent{\textbf{Max-min fairness selection.}} As the goal is to maximize the worst performance across all groups, this can be achieved using a \textit{greedy strategy}. Specifically, for each group $a \in \mathcal{A}$, we compare our model $h_a \circ f$ with the ERM baseline $h_{\text{erm}}$ and adopt the better-performing model.

\noindent{\textbf{Overall accuracy parity fairness selection.}}
To maximize overall performance while minimizing the performance gap among all groups without imposing harm, we treat this as a constrained combinatorial optimization problem that selects the optimal combination of models to achieve a smaller fairness gap without compromising overall performance. Given that the number of sensitive attributes is relatively small in fairness studies, it is feasible to find the exact solution using integer programming.
Specifically, for each group $a\in\mathcal{A}$, define a binary decision variable $ v_a \in\{0,1\}$, where {$ v_a = 1 $ if we choose our model $h_a\circ f$ for group $a$, and $ v_a = 0 $ }if we choose $h_{\text{erm}}$.  Let {${\alpha}_{\text{expert},a}$ and ${\alpha}_{\text{erm},a}$ be the model accuracy of group $a$ attained under expert $h_a\circ f$ and $h_{\text{erm}}$, respectively, then} the model accuracy of group $a$, denoted as $\alpha_a$, can be equivalently written as: 
\begin{equation}
{\alpha}_a = v_a \cdot {\alpha}_{\text{expert},a} + (1 - v_a) \cdot {\alpha}_{\text{erm},a}
\end{equation}
Let $p_a$ be the proportion of group $a$ in the entire population. We have the following constrained optimization:
\begin{equation}
\begin{aligned}
\min & \quad \Delta - \lambda \sum_{a \in \mathcal{A}} p_a \cdot \alpha_a \\
\text{subject to:} \quad & \alpha_a \geq \alpha_{\text{erm}, a} \quad \forall a \in \mathcal{A} \quad \text{(No-harm)} \\
                  & \Delta \geq \alpha_i - \alpha_j \quad \forall i, j \in \mathcal{A} \quad \text{(Fairness)} \\
                  & v_a \in \{0, 1\} \quad \forall a \in \mathcal{A}
\end{aligned}
\end{equation}

Here $\Delta$ represents the maximum difference between accuracies across groups, and $- \lambda \sum_a p_a \cdot {\alpha}_a $ directs the optimization to find a solution with higher accuracy.

\noindent \textbf{Feasibility Analysis.} In the worst case, considering the trivial solution where we choose the ERM model for all groups $\alpha_a = \alpha_{\text{erm},a}$, the accuracy constraint $\alpha_a \geq \alpha_{\text{erm},a}$ is trivially satisfied for all groups. The fairness constraint reduces to: $\Delta \geq \alpha_{\text{erm},i} - \alpha_{\text{erm},j}, \forall i, j \in \mathcal{A}$. Let \(\Delta_{\text{erm}}\) represent the maximum accuracy difference between any two groups under the ERM model, setting $\Delta = \Delta_{\text{erm}}$ satisfies this constraint. Therefore, the trivial solution is feasible. 

Consider a more general case where we select our models $h_a\circ f$ for some groups and the baseline $h_{\text{erm}}$ for others. The accuracy constraint guarantees that the accuracy of our model is at least as good as $h_{\text{erm}}$ for each group, while the fairness constraint minimizes the maximum accuracy difference across all groups. Although the trivial solution, where all groups use the ERM model, is feasible, we aim to explore better alternatives that satisfy both constraints and optimize the overall objective. This allows us to potentially achieve better fairness and performance across all groups.

\section{Experiments}
Experiments are run on a server with NVIDIA A5000 GPUs and AMD EPYC 7313 CPUs.  ResNet-18 \cite{he2016deep} serves as the backbone $f$, with separate linear classifiers $h$ for each group. Implementation details are in Appendix B.

\subsection{Datasets and Baselines}
We focus on medical datasets, where access to sensitive attributes is typically feasible and required by our method. Specifically, we evaluate FairSDE on three diagnostic tasks:  skin disease classification with \textbf{Ham10000} \citep{maron2019systematic}, chest X-ray interpretation with \textbf{MIMIC-CXR} \citep{johnson2019mimic}, and glaucoma detection using \textbf{Harvard-GF} \citep{10472539}. To provide a broader comparison, we also include two facial image datasets: \textbf{CelebA} \citep{liu2015faceattributes} and \textbf{UTKFace} \citep{zhifei2017cvpr}. Dataset details are provided in Appendix A.
As medical datasets often exhibit class imbalance, we report the AUC for disease prediction tasks and accuracy for facial recognition tasks.

We compare against the following baselines:
1) \textbf{ERM}: Serves as a no-harm baseline; methods must outperform ERM on each group to claim fairness without harm.  
2) \textbf{Decoupled Classifiers}: Assigns each group a classifier $h(X,A)$ trained with a shared feature extractor, adapted from \citet{ustun2019fairness}.  
3) \textbf{Adversarial Training}~\cite{zhang2018mitigating}: Employs an adversary to predict the sensitive attribute, which the model tries to obscure.  
4) \textbf{CFL}~\cite{shen2021contrastive}: Uses contrastive learning to align representations of samples with similar characteristics across groups.  
5) \textbf{FSCL}~\cite{park2022fair}: Encourages class-wise compactness while removing sensitive information from representations.  
6) \textbf{FIN}~\cite{10472539}: Uses group-specific feature normalization with learnable statistics to balance feature importance.  
7) \textbf{GroupDRO}~\cite{Sagawa2020Distributionally}: Applies group distributional robustness to mitigate spurious correlations.  
8) \textbf{FIS}~\cite{pang2025fairness}: Selects training samples based on utility and fairness influence to improve fairness without harm.

\subsection{Results}
The results for the medical and facial datasets are shown in Tables~\ref{table:comparison_medical} and \ref{table:comparison_face}. Experiments are repeated three times, and we report the average values. Additional results are provided in the Appendix, including:  \textbf{equalized odds}, \textbf{standard deviations}, and \textbf{ablation studies on each module}. We find that \textbf{FairSDE is the only method that consistently achieves fairness without harm}. Based on the results, we address the following research questions:

\begin{table*}[ht]
\small
\setlength{\tabcolsep}{1mm}
\centering
\begin{tabular}{lcc|ccccccc|cc}
\hline
\textbf{Dataset} & \textbf{Sensitive} & \textbf{Metric} & \textbf{ERM} & \textbf{Adversarial} & \textbf{CFL} & \textbf{FSCL+} & \textbf{FIN} & \textbf{GroupDRO}  & \textbf{FIS} & \textbf{Decoupled} & \textbf{FairSDE} \\
\hline
\multirow{6}{*}{Ham10000} & \multirow{3}{*}{Gender} & AUC & 84.07 & 83.70 $\downarrow$ & 83.67 $\downarrow$ & 82.18 $\downarrow$ & 83.33 $\downarrow$ & 84.45 & 84.55 & 83.60 $\downarrow$ & \textbf{84.75} \\
 &  & MF & 82.78 & 82.93 & 82.70  $\downarrow$ & 81.76  $\downarrow$ & 82.06  $\downarrow$ & 83.42 & 82.99 & 82.20  $\downarrow$ & \textbf{83.48} \\
 &  & Gap & 2.76 & 1.67 & 2.52 & \textbf{0.75} & 2.93 & 2.17 & 3.34 & 3.06 & 2.65 \\
\cline{2-12}
 & \multirow{3}{*}{Age} & AUC & 84.10 & 82.85 $\downarrow$ & 84.30 & 81.18 $\downarrow$ & 83.25 $\downarrow$  & 82.92 $\downarrow$ & 84.30 & 83.91 $\downarrow$ & \textbf{84.62} \\
 &  & MF & 75.13 & 72.88 $\downarrow$ & 74.56 $\downarrow$ & 70.67 $\downarrow$ & 73.78 $\downarrow$ & 74.48 $\downarrow$ & 75.17 & 73.73 $\downarrow$ & \textbf{76.03} \\
 &  & Gap & 13.05 & 15.26 & 14.37 & 15.82 & 13.30 & 12.23 & 13.34 & 14.68 & \textbf{12.17} \\
\hline
\multirow{6}{*}{Mimic-CXR} & \multirow{3}{*}{Gender} & AUC & 82.28 & 79.98 $\downarrow$ & 82.62 & \textbf{82.67} & 82.22 $\downarrow$ & 82.09 $\downarrow$ & 82.07 $\downarrow$ & 82.27 $\downarrow$ & 82.60 / 82.39 \\
 &  & MF & 81.53 & 79.42 $\downarrow$ & 81.81 & 81.71 & 81.54 & 81.43 $\downarrow$ & 81.33 $\downarrow$ & 81.62 & \textbf{81.92} \\
 &  & Gap & 1.54 & 1.34 & 1.62 & 1.94 & 1.36 & 1.33 & 1.52 & 1.41 & \textbf{1.33} \\
\cline{2-12}
 & \multirow{3}{*}{Race} & AUC & 82.28 & 81.56 $\downarrow$ & 82.50 & \textbf{82.66} & 82.23 $\downarrow$ & 81.81 $\downarrow$ & 82.23 $\downarrow$ & 82.32 & 82.44 \\
 &  & MF & 81.70 & 81.27 $\downarrow$ & 82.06 & 82.16 & 81.58 $\downarrow$ & 81.39 $\downarrow$ & 81.71 & 81.89 & \textbf{82.17} \\
 &  & Gap & 0.44 & 0.20 & 0.30 & 0.45 & 0.47 & 0.25 & 0.31 & \textbf{0.14} & 0.26 \\
\hline
\multirow{6}{*}{Harvard-GF} & \multirow{3}{*}{Gender} & AUC & 81.60 & 82.36 & 82.48 & 79.95 $\downarrow$ & 81.67 & 81.58 $\downarrow$ & 83.56 & \textbf{84.44} & 82.31 / 81.98\\
 &  & MF & 80.54 & 81.77 & 81.63 & 79.50 $\downarrow$ & 80.77 & 80.66  & 82.68 & \textbf{84.21} & 81.20 \\
 &  & Gap & 2.36 & 1.42 & 2.03 & 1.08 & 2.05 & 2.15 & 2.01 & \textbf{0.42} & 1.81 \\
\cline{2-12} 
 & \multirow{3}{*}{Race} & AUC & 82.59 & 82.54 $\downarrow$ & 82.61 & 79.34 $\downarrow$ & 82.23 $\downarrow$ & 82.63 & 83.42 & \textbf{84.81} & 82.83 \\
 &  & MF & 78.13 & 78.25 & 77.99 $\downarrow$ & 74.46 $\downarrow$ & 76.52 $\downarrow$ & 78.83 & 78.80 & \textbf{80.44} & 78.61 \\
 &  & Gap & 6.58 & 6.66 & 6.99 & 7.61 & 8.01 & 6.28 & 6.97 & 7.17 & \textbf{6.12} \\
\hline
\multirow{2}{*}{\shortstack{\textbf{Fairness} \\ \textbf{w/o Harm}}}
& \multicolumn{1}{c}{\textbf{Utility}} 
& \multicolumn{1}{c}{Win $\mid$ Loss} 
& - & 1$\mid$5 & 4$\mid$2 & 2$\mid$4 & 1$\mid$5 & 2$\mid$4 & 2$\mid$4 & 3$\mid$3 & \textbf{6$\mid$0} \\
& \multicolumn{1}{c}{\textbf{Fairness}} 
& \multicolumn{1}{c}{Win $\mid$ Loss} 
& - & 9$\mid$3 & 9$\mid$3 & 8$\mid$4 & 8$\mid$4 & 9$\mid$3 & 11$\mid$1 &  10/2 & \textbf{12$\mid$0} \\
\hline
\end{tabular}
\caption{Classification results on three medical datasets. The AUC metric is used to quantify classification performance. MF denotes max-min fairness, representing the lowest AUC across all groups. Overall accuracy parity is evaluated using the AUC metric, with the AUC gap between the most advantaged and disadvantaged groups reported as "Gap" in the table. When the \textit{greedy strategy} and \textit{Integer Programming} yield different solutions, both overall AUC values are reported for comparison.}
\label{table:comparison_medical}
\end{table*}

\begin{table*}[h!]
\small
\setlength{\tabcolsep}{1mm}
\centering
\renewcommand{\arraystretch}{1}

\begin{tabular}{lcc|ccccccc|cc}
\hline
\textbf{Dataset} & \textbf{Sensitive} & \textbf{Metric} & \textbf{ERM} & \textbf{Adversarial} & \textbf{CFL} & \textbf{FSCL+} & \textbf{FIN} & \textbf{GroupDRO} & \textbf{FIS} & \textbf{Decoupled} & \textbf{FairSDE} \\
\hline
\multirow{3}{*}{CelebA-Hair} & \multirow{3}{*}{Gender} & ACC & 81.28 & 79.67 $\downarrow$ & 81.49 & 81.69 & 81.13 $\downarrow$  & 79.63 $\downarrow$ &  80.14 $\downarrow$ & 81.40 & \textbf{82.54} \\
 &  & MF & 77.60 & 75.71  $\downarrow$ & 77.72 & 78.35 & 77.28 $\downarrow$ & 78.35 & 76.64 $\downarrow$  & 77.70 & \textbf{79.16} \\
 &  & Gap & 6.32 & 6.80 & 6.48 & {5.73} & 6.60 & \textbf{2.20}$\downarrow$ & 6.01  & 6.35 & 5.80 \\
\cline{2-12}
\multirow{3}{*}{CelebA-Smiling} & \multirow{3}{*}{Gender} & ACC & 91.87 & 90.32 $\downarrow$ & 91.88 & 91.76 $\downarrow$ & 91.73 $\downarrow$ & 91.50 $\downarrow$ &91.14 $\downarrow$ & 91.82 $\downarrow$ & \textbf{92.22} \\
 &  & MF & 90.68 & 89.19 $\downarrow$ & 90.64 $\downarrow$ & 90.72 & 90.53 $\downarrow$ & 90.98 &89.76 $\downarrow$  & 90.58 $\downarrow$ & \textbf{91.21} \\
 &  & Gap & 2.04 & 1.92 & 2.12 & 1.77 & 2.06  & \textbf{0.88} &2.36& 2.12  & {1.71} \\
\hline
\multirow{6}{*}{UTK} & \multirow{3}{*}{Gender} & ACC & 78.06 & 73.40  $\downarrow$ & 77.89  $\downarrow$ & 79.97 & 73.93  $\downarrow$ & 77.84 $\downarrow$  & 76.32 $\downarrow$ & 75.50  $\downarrow$ & \textbf{80.45} \\
 &  & MF & 76.52 & 71.95  $\downarrow$ & 76.21  $\downarrow$ & 78.62 & 70.89  $\downarrow$ & 76.83 & 74.86 $\downarrow$ & 72.58  $\downarrow$ & \textbf{79.52} \\
 &  & Gap & 3.29 & 3.27 & 3.60 & 2.87 & 6.49  & 2.15 & 3.11 & 6.23 & \textbf{1.98} \\
\cline{2-12}
 & \multirow{3}{*}{Race} & ACC & 81.65 & 78.39  $\downarrow$ & 82.84 & {83.62} & 78.86 $\downarrow$  & 82.83 & 81.61 $\downarrow$ & 81.39  $\downarrow$ & \textbf{84.17} \\
 &  & MF & 80.04 & 77.10  $\downarrow$ & 81.29 & {82.08} & 77.80  $\downarrow$  & 81.62 & 80.02 $\downarrow$ & 80.52 & \textbf{82.75} \\
 &  & Gap & 3.81 & 3.04 & 3.68 & 3.67 & 2.51  & 2.86  & 3.77 & \textbf{2.06} & 3.37 \\
\hline
\multirow{2}{*}{\shortstack{\textbf{Fairness} \\ \textbf{w/o Harm}}}
& \multicolumn{1}{c}{\textbf{Utility}} 
& \multicolumn{1}{c}{Win $\mid$ Loss} 
& - & 0$\mid$4 & 3$\mid$1 & 3$\mid$1 & 0$\mid$4 & 1$\mid$3 & 0$\mid$4 & 1$\mid$3 & \textbf{4$\mid$0} \\
& \multicolumn{1}{c}{\textbf{Fairness}} 
& \multicolumn{1}{c}{Win $\mid$ Loss} 
& - & 4$\mid$4 & 6$\mid$2 & 8$\mid$0 & 4$\mid$4 & 7$\mid$1 & 4$\mid$4 &  2$\mid$6 & \textbf{8$\mid$0} \\
\hline
\end{tabular}
\caption{Classification results on face datasets. The same reporting criteria as used in Table 1 are applied. For face datasets, accuracy (ACC) is used as the performance metric instead of AUC. Standard deviations are reported in Appendix D.}
\label{table:comparison_face}

\end{table*}

\smallskip
\noindent\textbf{Why Do We Need Fairness Without Harm?} As shown in Table \ref{table:comparison_medical}, methods like \textit{Adversarial} and \textit{FSCL+} significantly reduce AUC gap between male and female groups, suggesting improved fairness over baselines like \textit{ERM}. However, this fairness gain comes at a cost: \textit{FSCL+} reduces overall AUC by 1.89\% and lowers performance for the disadvantaged group by  1.02\%. A similar pattern emerges in glaucoma assessment on Harvard-GF data, where fairness methods compromise diagnostic accuracy. In healthcare, sacrificing accuracy for fairness is problematic and may have harmful consequences such as misdiagnosis or delayed treatment. Therefore, the concept of \textit{fairness without harm} is crucial in real world. It emphasizes the need to develop models that not only promote fairness across groups but also maintain high levels of accuracy and reliability for patients.

\smallskip
\noindent\textbf{Comparison and Discussion.} 
We conducted a comparative analysis of model performance across medical and face datasets, using a down arrow symbol~($\downarrow$) to indicate performance degradation relative to ERM. Note that \textbf{FairSDE is not designed to significantly outperform other methods in accuracy or fairness alone, but rather to ensure that fairness is consistently achieved without sacrificing performance}. Its expert selection mechanism may deliberately choose a model with slightly lower accuracy if it better satisfies the fairness without harm criterion. Among the baselines, \textit{FSCL+} performed well on face datasets, improving both fairness and accuracy over ERM. However, this improvement did not generalize to medical datasets, where \textit{FSCL+} yielded lower AUC than ERM. Other fairness methods showed similar performance degradation. \textbf{In contrast, FairSDE is the only method that consistently achieves fairness without harm, outperforming ERM across both AUC and ACC metrics without reducing performance.} This consistency highlights the strength of FairSDE, particularly in tasks like medical diagnosis, where it achieves fairness while simultaneously improving accuracy through the learned distinct representations and personalized classifiers. However, as in many studies, we assume that distributions are identical between validation and test sets. This assumption is critical, as our no-harm selection strategy may fail to perform as intended under a significant distribution shift.

\smallskip
\noindent\textbf{Quantitative Results of Existing Algorithms on Fairness Without Harm.} Next, we examine whether existing \textit{fairness without harm} algorithms can effectively handle more challenging datasets. For all datasets, we apply decoupled classifiers $h(X,A)$ based on sensitive attributes. Unlike the original study by \citet{ustun2019fairness}, we do not use preference guarantees to select attributes; instead, we evaluate decoupled classifiers independently for each sensitive attribute, using representations learned from the ERM model. 
As shown in Table \ref{table:comparison_medical}, decoupled classifiers perform well on datasets with balanced distributions of sensitive attributes, such as Harvard-GF, which has an equal number of samples across three racial groups and nearly equal representation of male and female groups. However, on more challenging datasets with imbalanced group distributions, these classifiers struggle to capture group-specific characteristics. For instance, on Ham10000, decoupled classifiers fail to outperform ERM on both gender and age attributes, leading to a trivial solution that defaults to ERM for all sensitive groups. This degraded performance highlights that trivially decoupled classifiers $h(X,A)$ are insufficient to achieve fairness without harm. We also evaluate another fairness-without-harm method, FIS, which uses active learning to selectively sample training data based on its fairness influence on the validation set. However, instead of enforcing no-harm constraint relative to the ERM model, FIS compares fairness improvements against previous training epochs. As a result, it may still underperform ERM in certain cases.

\begin{figure}[h]
        \centering
    \begin{subfigure}[t]{\linewidth}
        \centering
        \includegraphics[width=\linewidth]{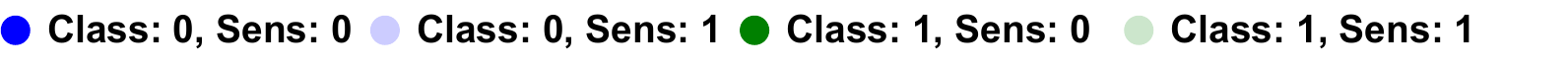}
    \end{subfigure}
        
    \begin{subfigure}[t]{0.43\linewidth}
        \centering
        \includegraphics[width=\linewidth]{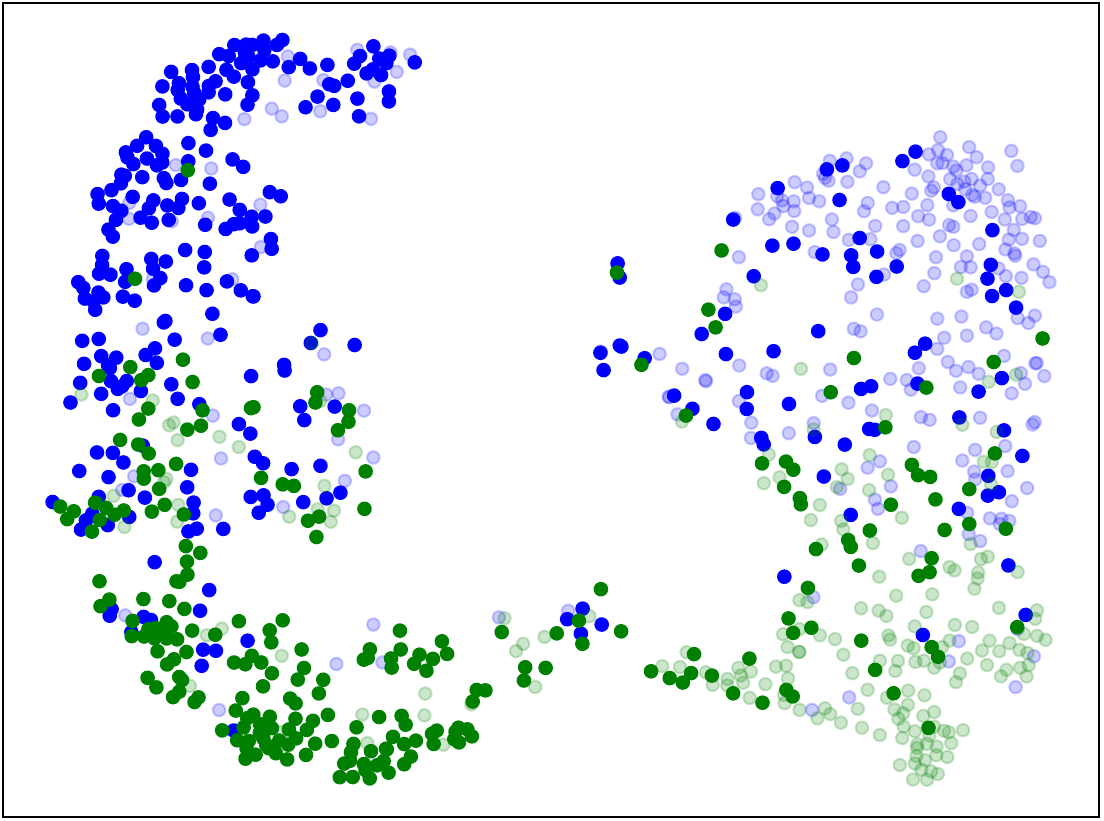}
        \caption{FairSDE}
        \label{fig:vis_ours}
    \end{subfigure}
    \begin{subfigure}[t]{0.43\linewidth}
        \centering
        \includegraphics[width=\linewidth]{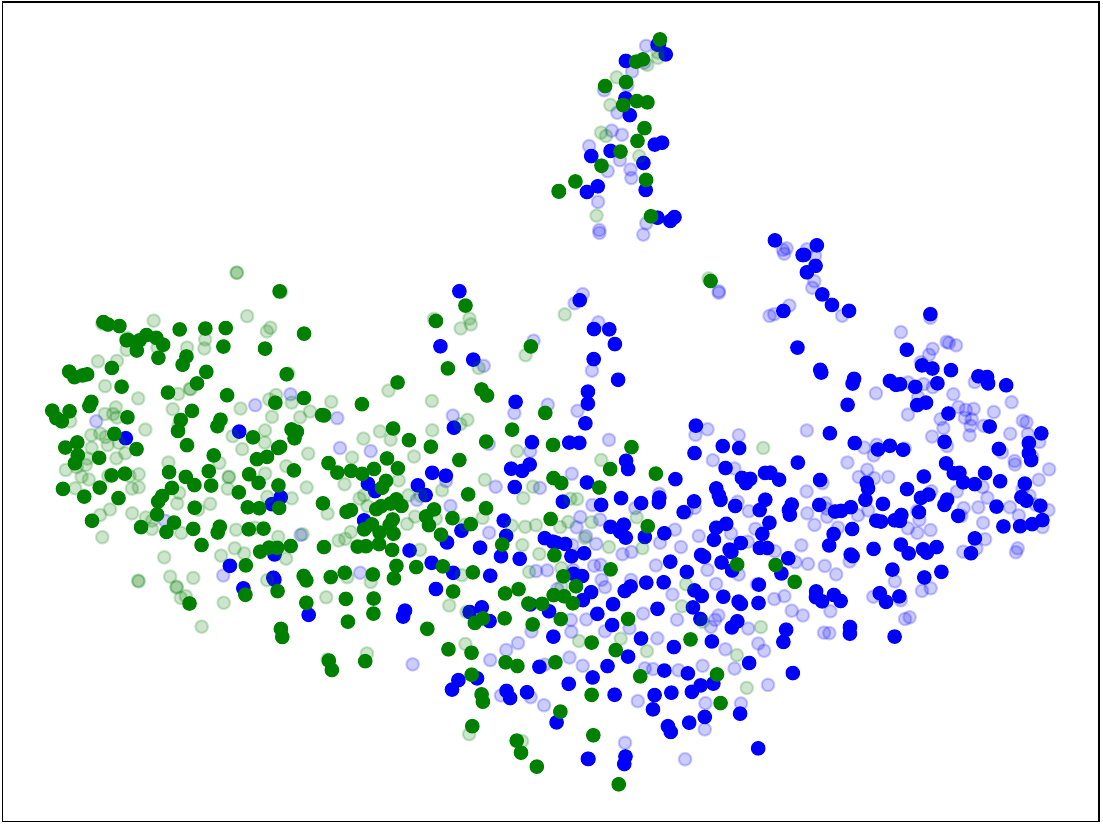}
        \caption{ERM}
        \label{fig:vis_erm}
    \end{subfigure}
    
    \begin{subfigure}[t]{0.43\linewidth}
        \centering
        \includegraphics[width=\linewidth]{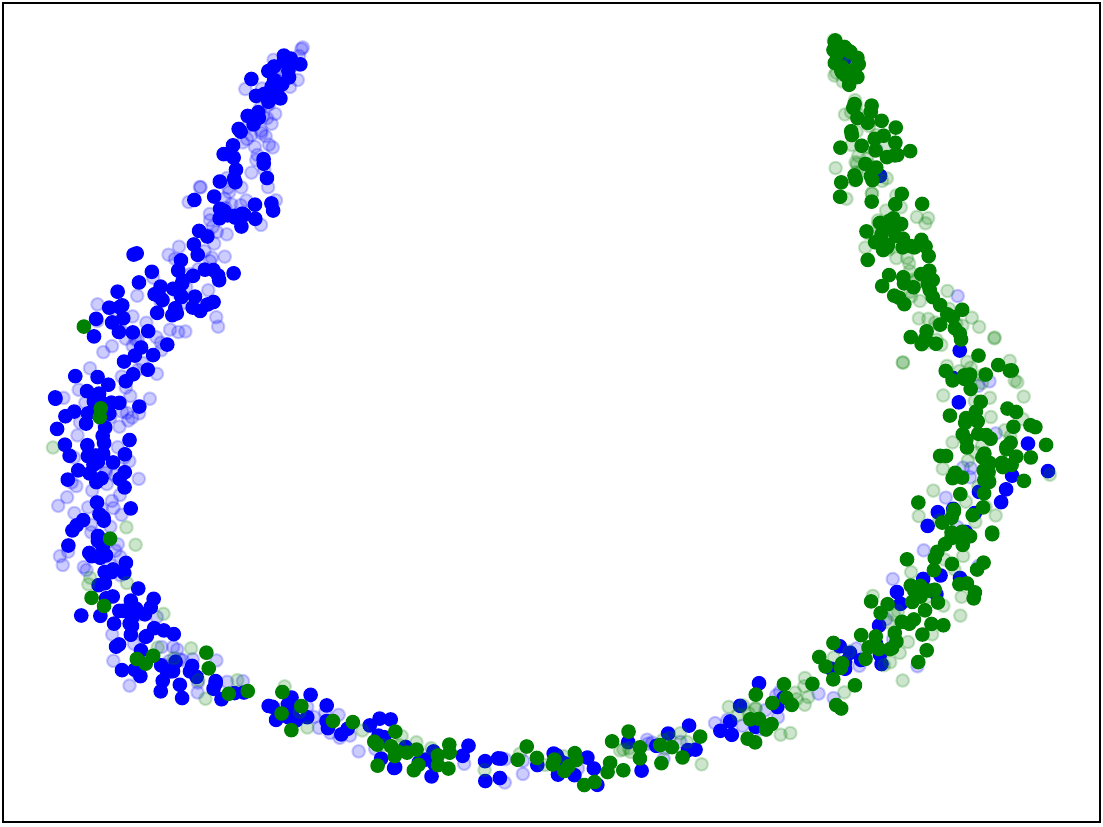}
        \caption{CFL}
    \end{subfigure}
    \begin{subfigure}[t]{0.43\linewidth}
        \centering
        \includegraphics[width=\linewidth]{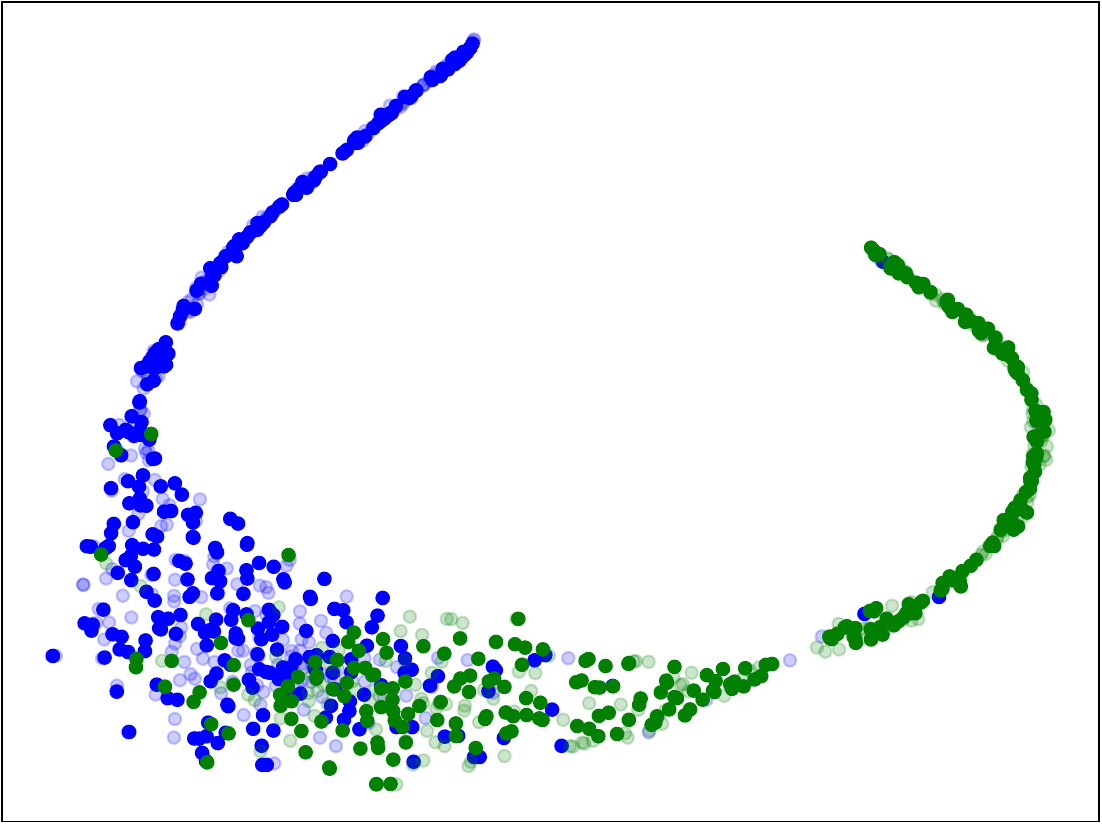}
        \caption{FSCL+}
    \end{subfigure}
    \caption{t-SNE visualizations of representations on the UTKFace. The two classes are represented in blue and green, while the \textit{white} group is depicted using darker circles and the \textit{non-white} group using lighter circles.}
    \label{fig:interesi}
\end{figure}

\noindent\textbf{How Do Decoupled Representations Perform on Images?}  Figure \ref{fig:interesi} presents t-SNE visualizations of the representations learned by FairSDE and other methods on UTKFace data, which includes two target classes (male and female) and two sensitive groups (white and non-white). As shown in Figure~\ref{fig:vis_ours}, FairSDE achieves a clear separation between classes and sensitive groups by learning from virtual centers and maintaining distinct group-wise  mean representations. This enables the model to better capture group-specific characteristics, leading to fairer and more accurate predictions. In contrast, Figure~\ref{fig:vis_erm} shows that ERM produces more entangled representations, with significant overlap across groups and classes.  Similarly, the other two methods tend to learn representations that are less sensitive to group distinctions and rely on a pooled classifier, as opposed to FairSDE.

From the MF results in Tables \ref{table:comparison_medical} and \ref{table:comparison_face}, we also observe that using demographic experts alone consistently improves the accuracy for the worst-performing group compared to the pooled classifier (ERM). This suggests that decoupling both representations and classifiers more effectively captures the unique patterns within disadvantaged groups. In contrast, applying decoupled classifiers without adapting the representations can degrade performance for disadvantaged groups, as evidenced by the poorer MF results of \textit{Decoupled} on the Ham10000. This highlights a key strength of FairSDE: decoupled representations are crucial for achieving fairness without compromising performance. By better modeling group-specific characteristics, FairSDE delivers fair and accurate predictions across all demographic groups.

\smallskip
\noindent\textbf{No-Harm Selection.} Using Harvard-GF as an example, Table~\ref{tab:noharmselect} presents no-harm selection results, highlighting the distinct advantages of the \textit{greedy (GS)}  and \textit{integer programming (IP)} strategies in improving fairness and overall AUC. The \textit{greedy} strategy aims to maximize the minimum group performance to ensure max-min fairness, as reflected by the increased average AUC for the worst-performing group (81.2) compared to ERM baseline (80.54). \textit{GS} often leads to better overall performance by selecting the optimal model for each group. In contrast, the \textit{IP} strategy seeks to minimize the performance gap between advantaged and disadvantaged groups by balancing selections between ERM and FairSDE. While this may slightly reduce overall performance, it effectively reduces disparities. For example, \textit{IP} frequently selects ERM model for Group 2 to avoid worsening performance gaps, recognizing that improving this group’s performance could cause unfairness. Table~\ref{tab:noharmselect} shows that \textit{IP} reduces the performance gap to 1.81, with no group underperforming relative to ERM, owing to the no-harm constraint in its optimization. This confirms that both strategies successfully fulfill our goal of fairness without harm.

\begin{table}[htp]
\centering
\small
\setlength{\tabcolsep}{1mm}
\renewcommand{\arraystretch}{1}

\begin{tabular}{|c|c|c|c|c|c|c|c|c|}
\hline
\multirow{2}{*}{\textbf{T}} & \multirow{2}{*}{\textbf{Select}} & \multicolumn{2}{c|}{\textbf{Group 1 AUC}} & \multicolumn{2}{c|}{\textbf{Group 2 AUC}} & \multirow{2}{*}{\textbf{AUC}} & \multirow{2}{*}{\textbf{MF}} & \multirow{2}{*}{\textbf{Gap}} \\ \cline{3-6}
 &  & \textbf{ERM} & \textbf{Ours} & \textbf{ERM} & \textbf{Ours} & & & \\ \hline
\multirow{4}{*}{\textbf{1}} 
 & \textbf{ERM} & 81.19 & - & 83.80 & - & 82.28 & 81.19 & 2.60 \\ \cline{2-9} 
 & \textbf{Ours} & - & 81.45 & - & 83.66 & 82.49 & 81.45 & 2.21 \\ \cline{2-9} 
 & \textbf{GS} & \multicolumn{2}{c|}{Ours: 81.45} & \multicolumn{2}{c|}{Ours: 83.66} & 82.49 & 81.45 & 2.21 \\ \cline{2-9} 
 & \textbf{IP} & \multicolumn{2}{c|}{Ours: 81.45} & \multicolumn{2}{c|}{ERM: 83.80} & 82.56 & 81.45 & 2.35 \\ \hline
 
\multirow{4}{*}{\textbf{2}} 
 & \textbf{ERM} & 79.98 & - & 82.90 & - & 81.32 & 79.98 & 2.92 \\ \cline{2-9} 
 & \textbf{Ours} & - & 80.38 & - & 83.83 & 81.99 & 80.38 & 3.45 \\ \cline{2-9} 
 & \textbf{GS} & \multicolumn{2}{c|}{ERM: 79.98} & \multicolumn{2}{c|}{Ours: 83.83} & 81.72 & 79.98 & 3.85 \\ \cline{2-9} 
 & \textbf{IP} & \multicolumn{2}{c|}{ERM: 79.98} & \multicolumn{2}{c|}{ERM: 82.90} & 81.32 & 79.98 & 2.92 \\ \hline
 
\multirow{4}{*}{\textbf{3}} 
 & \textbf{ERM} & 80.45 & - & 82.02 & - & 81.21 & 80.45 & 1.57 \\ \cline{2-9} 
 & \textbf{Ours} & - & 82.17 & - & 83.13 & 82.71 & 82.17 & 0.95 \\ \cline{2-9} 
 & \textbf{GS} & \multicolumn{2}{c|}{Ours: 82.17} & \multicolumn{2}{c|}{Ours: 83.13} & 82.71 & 82.17 & 0.95 \\ \cline{2-9} 
 & \textbf{IP} & \multicolumn{2}{c|}{Ours: 82.17} & \multicolumn{2}{c|}{ERM: 82.02} & 82.06 & 82.02 & 0.15 \\ \hline
\multirow{3}{*}{\textbf{Ave}} 
 & \textbf{ERM} & $80.54 $ & - & $82.91 $ & - & $81.60 $ & $80.54 $ & $2.36 $ \\ \cline{2-9} 
 & \textbf{GS} & \multicolumn{2}{c|}{$81.20 $} & \multicolumn{2}{c|}{ $83.54 $ } & $82.31$ & $81.20$ & $2.34 $ \\ \cline{2-9} 
 & \textbf{IP} & \multicolumn{2}{c|}{$81.20$} & \multicolumn{2}{c|}{$82.91 $} & $81.98$  & $81.15 $  & $1.81 $\\ \hline

\end{tabular}
\caption{Selection examples report the group AUC on Harvard-GF, along with overall AUC, max-min fairness, and AUC gap over three trials (T).  We also show the selection results from the \textit{greedy~(GS)}, which maximizes max-min fairness, and \textit{integer programming~(IP)}, which targets overall accuracy parity. Note that group selection is based on the validation set. As a result, in Trial 2, the \textit{greedy strategy} selected ERM for Group 1, even though FairSDE achieved better performance for this group on the test set. }
\label{tab:noharmselect}

\end{table}

\section{Conclusion}

This paper studied the fairness without harm problem and showed that existing methods often improve fairness at the cost of group-specific performance. To address this, we proposed FairSDE, which learns distinct demographic representations and employs personalized experts to ensure fairness without degrading any group’s performance. By selectively applying these experts, we ensure fairness constraints are met without compromising any group’s performance.  Experiments on multiple real datasets validate our approach. Despite potential concerns, using sensitive attributes is practical in medical contexts where they are routinely collected.

\section*{Acknowledgements}
This work was funded in part by the National Science Foundation under award number IIS-2145625, IIS-2202699, and IIS-2416895, and by the National Institutes of Health under awards number R01AI188576.

\bibliography{aaai}

\newpage
\onecolumn
\appendix
\setcounter{secnumdepth}{2}

\section{Dataset Details}
\label{app:datasets}
\subsection{Ham10000}
Ham10000 consists of dermatoscopic images of pigmented skin lesions. We reclassified the 7 diagnostic categories into two broad labels: ``benign'' and ``malignant,'' following \cite{maron2019systematic}. The ``benign'' category includes basal cell carcinoma (bcc), benign keratosis-like lesions (such as solar lentigines, seborrheic keratoses, and lichen-planus-like keratoses, bkl), dermatofibroma (df), melanocytic nevi (nv), and vascular lesions (including angiomas, angiokeratomas, pyogenic granulomas, and hemorrhages, vasc). The ``malignant'' category comprises actinic keratoses and intraepithelial carcinoma/Bowen’s disease (akiec), as well as melanoma (mel). Images missing recorded sensitive attributes were excluded from the dataset. HAM10000 dataset includes two sensitive attributes: age and sex. We binarized the age attribute into two categories: ``young'' and ``old.'' Individuals aged 0-60 years were classified as ``young,'' while those aged 60 years and above were classified as ``old.'' Images missing recorded sensitive attributes, including age or sex, were excluded from the dataset. We split the dataset into training, testing, and validation sets with a ratio of 8:1:1, respectively. 

\subsection{MIMIC-CXR} MIMIC-CXR is a large collection of chest radiographs (X-rays) and corresponding radiology reports. We utilized a subset of the MIMIC-CXR dataset, consisting of 51,450 images. For this study, we used the ``No Finding'' attribute as the classification target, which indicates that no symptoms were found in the images. Two sensitive attributes from the MIMIC-CXR dataset were considered: sex and race. To balance the number of images in different groups, we binarized the race attribute into two categories: ``white'' and ``non-white.'' Additionally, we discarded samples where the race attribute was labeled as ``OTHER.'' The dataset was also split into training, testing, and validation sets with a ratio of 8:1:1.

\subsection{Harvard-GF}
The Harvard-GF dataset consists of glaucoma data from 3,300 patients with a relatively balanced distribution. Each patient is associated with an OCT RNFLT map, totaling 3,300 OCT RNFLT maps. The dataset also includes additional information such as visual field data, patient age, sex, race, and glaucoma labels. The dataset is split into 2,100 samples for training, 300 samples for validation, and 900 samples for testing. In this study, we focus on the 2D OCT RNFLT images. The male attribute is used as the sensitive attribute, denoted as Gender in our experiments, and the race attribute is also utilized. Unlike other datasets, the Harvard-GF dataset provides an equal number of samples for three racial groups: Asian, Black, and White. Therefore, we do not binarize the race attribute and retain all three racial categories in our experiments.

\subsection{CelebA}
The CelebFaces Attributes Dataset (CelebA) is a large-scale face attributes dataset containing more than 200,000 celebrity images, each annotated with 40 different attributes. For this study, we use the "straight hair" and "smiling" attributes as the classification target. The gender attribute is used as a sensitive attribute in our analysis. The dataset is split into training, validation, and testing sets with a ratio of 8:1:1, ensuring an effective distribution of data for model training, evaluation, and validation.

\subsection{UTKFace}
The UTKFace dataset consists of over 20,000 face images, each annotated with age, gender, and ethnicity. We also binarize the race attribute into two categories: "white" and "non-white". Then we address two tasks: (1) gender recognition, where race is used as the sensitive attribute, and (2) race recognition, where gender is used as the sensitive attribute. The dataset was split into training, validation, and testing sets following the same 8:1:1 ratio as used for CelebA.

\begin{table}[h]
\centering

\renewcommand{\arraystretch}{1.2}
\begin{tabular}{lp{0.6\linewidth}}

\hline
\textbf{Dataset}              & \textbf{Download Link} \\ \hline
HAM10000                      & \url{https://dataverse.harvard.edu/dataset.xhtml?persistentId=doi:10.7910/DVN/DBW86T} \\ \hline
MIMIC-CXR                     & \url{https://physionet.org/content/mimic-cxr/} \\ \hline
Harvard-GF                    & \url{https://github.com/Harvard-Ophthalmology-AI-Lab/Harvard-GF} \\ \hline
CelebA                        & \url{https://mmlab.ie.cuhk.edu.hk/projects/CelebA.html} \\ \hline
UTKFace                       & \url{https://susanqq.github.io/UTKFace/} \\ \hline
\end{tabular}
\caption{Download links for the datasets used in this study.}
\label{table:datasets}
\end{table}

\section{Implementation Details}
\label{app:implementationdetails}

We present our algorithm in Algorithm \ref{alg:reprelearning}. All experiments were conducted on a server equipped with multiple NVIDIA A5000 GPUs, dual AMD EPYC 7313 CPUs, and 256GB of memory. The code is implemented with Python 3.8 and PyTorch 1.13.0 on Ubuntu 20.04. We utilized the SGD optimizer with a momentum of 0.9 and performed a grid search to fine-tune hyperparameters across all datasets. ResNet-18 was trained from scratch for 60 epochs across all methods. Separate linear classifiers $h$ are used for each demographic group. For some complex settings with multiple sensitive attributes, this adds only a few thousand parameters on top of a ResNet‑18‑scale backbone (for linear heads on 512-dimensional features with $|\mathcal{Y}|$ classes and $|\mathcal{A}|$ groups, the added parameters are $|\mathcal{A}|\times(512\times|\mathcal{Y}| + |\mathcal{Y}|)$; for example, with two groups ($|\mathcal{A}|=2$) and two classes ($|\mathcal{Y}|=2$), this results in $2\times(512\times2 + 2)=2052$ additional parameters). The virtual centers are initialized with the Kaiming Uniform.

The coefficients $\lambda_1$, $\lambda_2$, and $\lambda_3$ were selected from the range $[1e-2, 1]$, which is dataset-dependent; the range $[10^{-2}, 10^{-1}]$ works for most datasets.  Following hyperparameter tuning from the range $[1e-3, 1]$ for learning rates, the learning rate was set to $0.05$ for the UTKFace dataset and $0.01$ for the remaining datasets. A scheduler was applied to reduce learning rates by a factor of 0.9 after each epoch. A batch size of 256 was used for all experiments. For the baseline \textit{FSCL+}, we follow their implementation that first updates the representation function using contrastive learning, and then updates the classifier during the final 10 epochs. The FIS method was originally designed for an active learning setting, where a portion of the dataset remains unlabeled. To adapt it to the supervised learning setting, we make the unlabeled ratio as a hyperparameter for each dataset and employ ground-truth labels for influence-guided selection. Images from the Ham10000 and MIMIC-CXR datasets were resized to $224\times224$, while images from the CelebA and UTKFace datasets were resized to $64\times64$.

\begin{algorithm}
\caption{Diversity Representation Learning}
\begin{algorithmic}[1]

\REQUIRE Dataset $\mathcal{S}$, discriminator $D$, representation function $f$, virtual centers $\{V_{a,y}\}_{a \in \mathcal{A}, y \in \mathcal{Y} }$, group-specific classifiers $\{h_a\}_{a \in \mathcal{A}}$, coefficients $\lambda_1, \lambda_2, \lambda_3$, learning rate $\eta$.
 
\FOR{ each batch $ B \gets \{({x_a}, {a_i}, y_i)\}_{i=1}^{|B|} \subset \mathcal{S} $}
    \STATE $ \displaystyle \mathcal{L}_{\text{disc}} \gets - \sum_{i=1}^{|B|} \sum_{a \in \mathcal{A}} \mathbb{I}(a_i = a) \log D(f(x_i))_a$
    \STATE $\displaystyle \mathcal{L}_{\text{virt}} \gets  - \sum_{i}^{|B|} \sum_{a \in \mathcal{A}} \log\left(\frac{\exp(d(V_{a,y_i}, f(x_i)))}{\sum_{y' \in \mathcal{Y}} \exp(d(V_{a,y'}, f(x_i)))}\right)$
    
    \FOR{each sample $x_i$ in $B$}
        \STATE $x'_i \gets \text{RandomSample}(\{x_j \mid y_j = y_i, a_j = a_i, \, j \neq i \})$
        \STATE $x_i^{-} \gets \text{RandomSample}(\{x_j \mid y_j \neq y_i, a_j \neq a_i, \, j \neq i \})$
    
    \ENDFOR
    \STATE {$ \displaystyle \mathcal{L}_{\text{div}} = - \sum_{i=1}^{|B|} \log  \frac{\exp\left(f(x_i) \cdot f(x_i')\right) + \exp\left(d(V_{a_i,y_i}, f(x_i))\right)}{\exp\left(f(x_i) \cdot f(x_i^{-})\right) + \sum_{a \neq a_i, y \neq y_i } \exp\left(d(V_{a,y}, f(x_i))\right)}$}
    
    \FOR{ $a$ in $\mathcal{A}$}
    \STATE $\mathcal{L} \gets \sum_{i:a_i=a}^{|B|} \text{CrossEntropy}(h_a(f(x_i),y_i))$
    \ENDFOR
    \STATE $D \gets D - \eta  \nabla_D  { \lambda_1 \mathcal{L}_{\text{disc}}}$
    \STATE $V_{a,y} \gets V_{a,y} - \eta  \nabla_{V_{a,y}} (\lambda_2\mathcal{L}_{\text{virt}}+\lambda_3\mathcal{L}_{\text{div}} ), \forall a \in \mathcal{A}, y \in \mathcal{Y}$
    
    \STATE $ f \gets f - \eta \nabla_f (\mathcal{L}+\lambda_1\mathcal{L}_{\text{disc}}+\lambda_2\mathcal{L}_{\text{virt}}+\lambda_3\mathcal{L}_{\text{div}})$
    \STATE $h_a \gets h_a - \eta \nabla_{h_a} \mathcal{L}, \forall a \in \mathcal{A}$
\ENDFOR

\end{algorithmic}
\label{alg:reprelearning}
\end{algorithm}

\section{Addition Experiments}

\noindent\textbf{Equalized Odds: }In addition to max-min fairness and overall accuracy parity discussed in the main text, here we report the results with equalized odds: 
\begin{equation*}
\mathbb{P}(\hat{Y} = 1 \mid A = 0, Y = y) = \mathbb{P}(\hat{Y} = 1 \mid A = 1, Y = y), 
\quad y \in \{0, 1\}
\end{equation*}

To quantify equalized odds, we consider the following metric:
\begin{equation*}
1 - \frac{1}{2} \Big( 
\left| \mathbb{P}(\hat{Y} = 1 \mid A = 0, Y = 0) - \mathbb{P}(\hat{Y} = 1 \mid A = 1, Y = 0) \right| 
+ 
\left| \mathbb{P}(\hat{Y} = 1 \mid A = 0, Y = 1) - \mathbb{P}(\hat{Y} = 1 \mid A = 1, Y = 1) \right| 
\Big)
\end{equation*}

The results, reported in Table \ref{table:eoqq} (results are from models in Table 1-2), show FairSDE consistently improves equalized odds without compromising the performance compared with ERM. However, as FairSDE aims to achieve fairness by ensuring no harm to any group, rather than solely maximizing any single fairness metric like other bias mitigation algorithms,  it does not guarantee outperformance against specialized algorithms tailored to specific fairness criteria.  Instead, FairSDE excels in its versatility, effectively addressing fairness concerns across diverse scenarios without sacrificing overall predictive quality.

\begin{table*}[h]
\centering
\renewcommand{\arraystretch}{1.2}

\begin{tabular}{lccccccc}
\hline
\textbf{Method}  & \multicolumn{2}{c}{\textbf{Ham10000}} & \multicolumn{2}{c}{\textbf{Mimic-CXR}} & \multicolumn{2}{c}{\textbf{Harvard-GF}} \\
\hline
\textbf{Sensitive Attribute} & Gender & Age & Gender & Race & Gender & Race \\
\hline
ERM (AUC)      & $84.07 \pm 0.24$ & $84.10 \pm 0.24$ & $82.28 \pm 0.04$ & $82.28 \pm 0.04$ & $81.60 \pm 0.48$ & $82.59 \pm 0.42$ \\
FairSDE (AUC)   & ${84.75} \pm 0.22$ & ${84.62} \pm 0.40$ & $82.60 \pm 0.16$ & $82.44 \pm 0.20$ & $82.31 \pm 0.43$ & $82.83 \pm 0.80$ \\
ERM (EO)      & $96.26 \pm 1.06$ & $89.64 \pm 0.85$ & $96.00 \pm 0.51$ & $92.99 \pm 0.54$ & $94.51 \pm 0.68$ & $92.10 \pm 1.84$ \\
FairSDE (EO)   & $97.86 \pm 0.65$ & $89.64 \pm 0.85$ & $96.64 \pm 1.05$ & $93.31 \pm 0.62$ & $97.55 \pm 1.01$ & $94.79 \pm 1.94$ \\
\hline
\end{tabular}

\begin{tabular}{lccccc}
\hline
\textbf{Method}   &  \textbf{CelebA-Hair} & \textbf{CelebA-Smiling} & \multicolumn{2}{c}{\textbf{UTK}} \\
\hline
\textbf{Sensitive Attribute}  & Gender & Gender & Gender & Race \\
\hline
ERM (ACC)       & $81.28 \pm 0.05$ & $91.87 \pm 0.04$ &  $78.06 \pm 0.55$ & $81.65 \pm 0.45$ \\
FairSDE (ACC)   & ${82.54} \pm 0.15$  & ${92.22} \pm 0.07$ & ${80.45} \pm 0.34$ & ${84.17} \pm 0.37$ \\
ERM (EO)       & $93.71 \pm 0.14$ & $96.11 \pm 0.16$ & $95.89 \pm 0.69$ & $96.56 \pm 0.53$ \\
FairSDE (EO)   & $95.39 \pm 1.15$ & $96.32 \pm 0.13$ & $97.97 \pm 0.72$ & $96.76 \pm 0.66$ \\
\hline
\end{tabular}
\caption{Equalized odds and utility results.}
\label{table:eoqq}
\end{table*}

\noindent\textbf{Ablation Study: }We conduct an ablation study to analyze the importance of each module in our method. Specifically, we investigate the impact of the discriminator and the impact of diversity learning. We first consider removing the discriminator from our model and only use virtual centers to learn the diversity representation (denoted as \textit{w/o D}). In contrast, we only use the discriminator and decoupled classifiers without using virtual centers to study the impact of diversity learning (denoted as \textit{w/o V}). We further ablate the randomly pairwise alignment and similarity penalization by removing $\mathcal{L}_{div}$ from optimization to show the importance of reducing variance by compacting space, denoting it as \textit{w/o C}. Here, we use Gender as the sensitive attribute and evaluate FairSDE and these variants, the results are reported in Table \ref{table:ablation}. The results indicate that diversity learning is the most important component of the method. Without it, the model experiences a significant increase in disparity and the overall performance degrades to the ERM as no-harm selection finds the trivial solution.

\begin{table*}[h!]
\small
\centering
\renewcommand{\arraystretch}{1.1}
\setlength{\tabcolsep}{1mm}
\begin{tabular}{lccccc}
\hline
\textbf{Dataset}  & \textbf{Metric} & \textbf{ w/o D} & \textbf{ {w/o V}}  & \textbf{ {w/o C}} & \textbf{FairSDE} \\
\hline
\multirow{3}{*}{MIMIC-CXR} & AUC   &  $82.46 \pm 0.09$ / $82.28 \pm 0.04$   & $82.28 \pm 0.04$ & $82.45 \pm 0.12$ & ${82.60} \pm 0.16$ / $82.39 \pm 0.15$ \\
 &  MF &  $82.18 \pm 0.05$ &   $81.55 \pm 0.08$  & $81.73 \pm 0.22$ & ${81.92} \pm 0.32$ \\
 &  Gap &  $1.54 \pm 0.22$ &   $1.51 \pm 0.26$  &  $1.48 \pm 0.30$ &  ${1.33} \pm 0.35$ \\
\hline
\multirow{3}{*}{UTK}  & ACC & $83.53 \pm 0.04$ &  $82.52 \pm 0.37$ / $82.14 \pm 0.28$ & $84.00 \pm 0.19$ & $84.17 \pm 0.37$ \\
 &  MF  & $81.86 \pm 0.51$ &  $80.69 \pm 0.60$    & $82.05 \pm 0.15$ & $82.75 \pm 0.19$ \\
 &  Gap & $3.96 \pm 1.14$ & $3.43 \pm 1.38$ & $4.62 \pm 0.12$ & $3.37 \pm 0.73$  \\
\hline
\end{tabular}
\caption{Ablation study on model components: We evaluate three variants of FairSDE, including a version without the discriminator (\textit{w/o D}), a version without virtual centers (\textit{w/o V}), and a version without variance reduction in the representation distribution (\textit{w/o C}).  }
\label{table:ablation}
\end{table*}

\begin{figure}[h]
    \centering
    \includegraphics[width=0.6\textwidth]{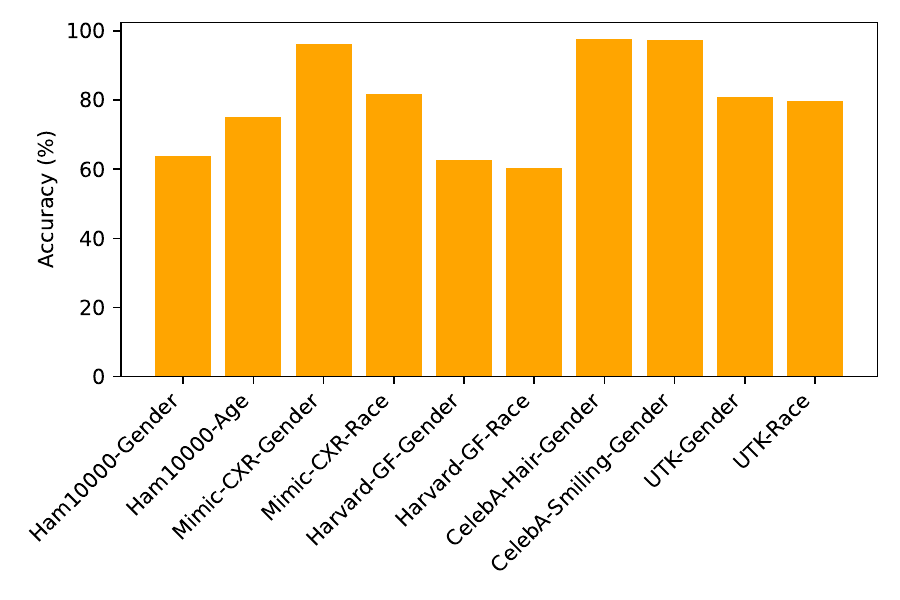}
    \caption{Group Discriminator Accuracy across Datasets.}
    \label{fig:dis_acc}
\end{figure}

\noindent\textbf{Group Discriminator Performance: } We report the group discriminator’s accuracy at the final stage as a quantitative separability check. As the figure \ref{fig:dis_acc} shows, the representations are indeed separable in most cases. However, for some datasets like Harvard-GF, focused on eye disease, it would be challenging to learn perfect separability due to the nature of certain medical images. However, in our method, we explicitly encourage, but do not require perfect separability.

\begin{figure}[h!]
    \centering
    \includegraphics[width=0.27\textwidth]{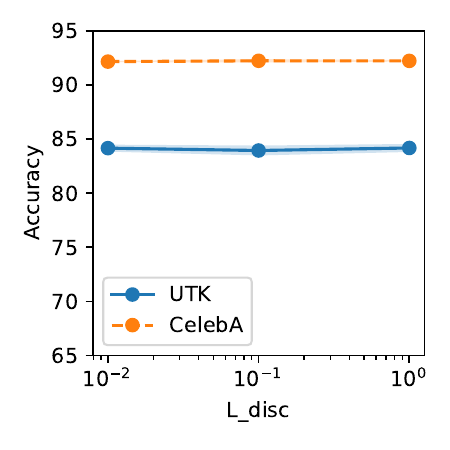}
    \includegraphics[width=0.27\textwidth]{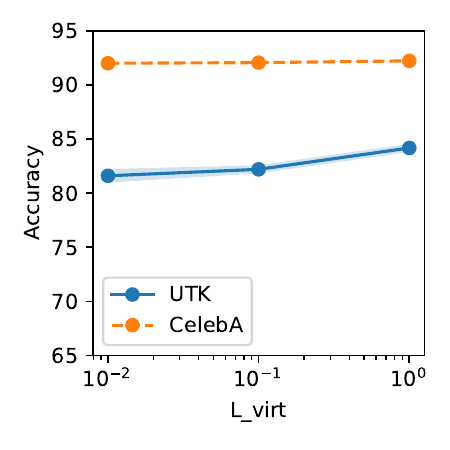}
    \includegraphics[width=0.27\textwidth]{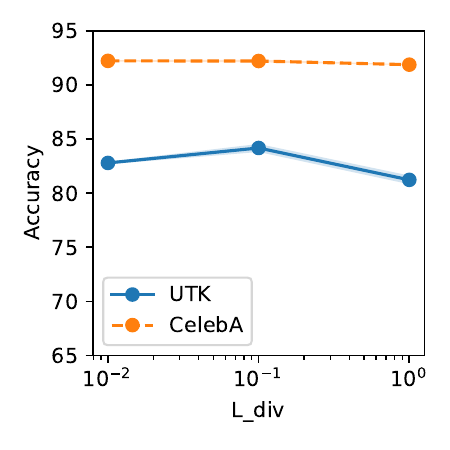}
    \caption{The average accuracy across two datasets with varying the hyperparameters.}
    \label{fig:sensitivity_study}
\end{figure}

\noindent\textbf{Sensitivity Study: } To further evaluate our method, we conduct a sensitivity study to analyze the impact of hyperparameter selection on model performance. Specifically, we examine the effect of varying the key hyperparameters $\mathcal{L}_{disc}$, $\mathcal{L}_{virt}$, and $\mathcal{L}_{div}$. The experiments are conducted on CelebA and UTKFace datasets, and the results are reported in Figure \ref{fig:sensitivity_study}. The results demonstrate that our method is not highly sensitive to the choice of hyperparameters, however, careful tuning of them for each dataset can yield noticeable performance improvements.

\section{Supplementary Results}

In this section, we report the standard deviations of results included in the main content in Table \ref{table:comparison_medical_full} and Table \ref{table:comparison_face_full}.

\begin{table*}[ht!]
\setlength{\tabcolsep}{1mm}
\centering
\renewcommand{\arraystretch}{1.2}
\resizebox{\linewidth}{!}{

\begin{tabular}{lcc|ccccccc|cc}
\hline
\textbf{Dataset} & \textbf{Sensitive} & \textbf{Metric} & \textbf{ERM} & \textbf{Adversarial} & \textbf{CFL} & \textbf{FSCL+} & \textbf{FIN}& \textbf{GroupDRO}& \textbf{FIS} & \textbf{Decoupled} & \textbf{FairSDE} \\
\hline
\multirow{6}{*}{Ham10000} & \multirow{3}{*}{Gender} & AUC & $84.07 \pm 0.24$ & $83.70 \pm 0.18 \downarrow$ & $83.67 \pm 0.10 \downarrow$ & $82.18 \pm 0.38 \downarrow$ & $83.33 \pm 0.41 \downarrow$ & $84.45 \pm 0.19$ & $84.55 \pm 0.31$ & $83.60 \pm 0.12 \downarrow$ & $\textbf{84.75} \pm 0.22$ \\
 &  & MF & $82.78 \pm 0.26$ & $82.93 \pm 0.54$ & $82.70 \pm 0.08 \downarrow$ & $81.76 \pm 0.35 \downarrow$ & $82.06 \pm 0.43 \downarrow$ & $83.42 \pm 0.31$ & $82.99 \pm 0.66$ & $82.20 \pm 0.32 \downarrow$ & $\textbf{83.48} \pm 0.18$ \\
 &  & Gap & $2.76 \pm 0.23$ & $1.67 \pm 0.84$ & $2.52 \pm 0.35$ & $\textbf{0.75} \pm 0.16$ & $2.93 \pm 0.87$ & $2.17 \pm 0.49$ & $3.34 \pm 0.86$ & $3.06 \pm 0.49$ & $2.65 \pm 0.47$ \\
\cline{2-12}
 & \multirow{3}{*}{Age} & AUC & $84.10 \pm 0.24$ & $82.85 \pm 0.20 \downarrow$ & $84.30 \pm 0.20$ & $81.18 \pm 1.05 \downarrow$ & $83.25 \pm 0.43$ & $82.92 \pm 0.54 \downarrow$ & $84.30 \pm 0.11$ & $83.91 \pm 0.21 \downarrow$ & $\textbf{84.62} \pm 0.40$ \\
 &  & MF & $75.13 \pm 0.14$ & $72.88 \pm 0.66 \downarrow$ & $74.56 \pm 0.29 \downarrow$ & $70.67 \pm 1.62 \downarrow$ & $73.78 \pm 0.79 \downarrow$ & $74.48 \pm 1.15 \downarrow$ & $75.17 \pm 0.48$ & $73.73 \pm 0.08 \downarrow$ & $\textbf{76.03} \pm 0.20$ \\
 &  & Gap & $13.05 \pm 0.65$ & $15.26 \pm 0.81$ & $14.37 \pm 0.19$ & $15.82 \pm 0.86$ & $13.30 \pm 1.21$ & $12.23 \pm 1.99 $ & $13.34 \pm 0.67$ & $14.68 \pm 0.37$ & $\textbf{12.17} \pm 0.34$ \\
\hline
\multirow{6}{*}{Mimic-CXR} & \multirow{3}{*}{Gender} & AUC & $82.28 \pm 0.04$ & $79.98 \pm 0.16 \downarrow$ & $82.62 \pm 0.01$ & $\textbf{82.67} \pm 0.13$ & $82.22 \pm 0.08 \downarrow$ & $82.09 \pm 0.13 \downarrow$ & $82.07 \pm 0.18 \downarrow$ & $82.27 \pm 0.18 \downarrow$ & $82.60 \pm 0.16$ \\
 &  & MF & $81.53 \pm 0.07$ & $79.42 \pm 0.34 \downarrow$ & $81.81 \pm 0.02$ & $81.71 \pm 0.10$ & $81.54 \pm 0.24$ & $81.43 \pm 0.26 \downarrow$ & $81.33 \pm 0.20 \downarrow$ & $81.62 \pm 0.14$ & $\textbf{81.92} \pm 0.32$ \\
 &  & Gap & $1.54 \pm 0.22$ & $1.34 \pm 0.45$ & $1.62 \pm 0.04$ & $1.94 \pm 0.14$ & $1.36 \pm 0.31$ & $1.33 \pm 0.27$ & $1.52 \pm 0.28$ & $1.41 \pm 0.19$ & $\textbf{1.33} \pm 0.35$ \\
\cline{2-12}
 & \multirow{3}{*}{Race} & AUC & $82.28 \pm 0.04$ & $81.56 \pm 0.24 \downarrow$ & $82.50 \pm 0.15$ & $\textbf{82.66} \pm 0.06$ & $82.23 \pm 0.08 \downarrow$ & $81.81 \pm 0.06 \downarrow$ & $82.23 \pm 0.08 \downarrow$ & $82.32 \pm 0.14$ & $82.44 \pm 0.20$ \\
 &  & MF & $81.70 \pm 0.26$ & $81.27 \pm 0.23 \downarrow$ & $82.06 \pm 0.11$ & $82.16 \pm 0.21$ & $81.58 \pm 0.32 \downarrow$ & $81.39 \pm 0.12 \downarrow$ & $81.71 \pm 0.12$ & $81.89 \pm 0.17$ & $\textbf{82.17} \pm 0.14$ \\
 &  & Gap & $0.44 \pm 0.32$ & $0.20 \pm 0.09$ & $0.30 \pm 0.27$ & $0.45 \pm 0.08$ & $0.47 \pm 0.13$ & $0.25 \pm 0.04$ & $0.31 \pm 0.19$ & $\textbf{0.14} \pm 0.11$ & $0.26 \pm 0.17$ \\
\hline
\multirow{6}{*}{Harvard-GF} & \multirow{3}{*}{Gender} & AUC & $81.60 \pm 0.48$ & $82.36 \pm 0.61$ & $82.48 \pm 0.14$ & $79.95 \pm 0.40 \downarrow$ & $81.67 \pm 0.69$ & $81.58 \pm 0.49 \downarrow$ & $83.56 \pm 0.71$ & $\textbf{84.44} \pm 0.50$ & $82.31 \pm 0.43 / 81.98 \pm 0.51$ \\
 &  & MF & $80.54 \pm 0.50$ & $81.77 \pm 0.73$ & $81.63 \pm 0.12$ & $79.50 \pm 0.17 \downarrow$ & $80.77 \pm 0.84$ & $80.66 \pm 0.54$ & $82.68 \pm 0.47$ & $\textbf{84.21} \pm 0.52$ & $81.20 \pm 0.91$ \\
 &  & Gap & $2.36 \pm 0.58$ & $1.42 \pm 0.60$ & $2.03 \pm 0.34$ & $1.08 \pm 0.47$ & $2.05 \pm 0.38$ & $2.15 \pm 0.28$ & $2.01 \pm 0.64$ & $\textbf{0.42} \pm 0.17$ & $1.81 \pm 1.19$ \\
\cline{2-12} 
 & \multirow{3}{*}{Race} & AUC & $82.59 \pm 0.42$ & $82.54 \pm 0.15 \downarrow$ & $82.61 \pm 0.15$ & $79.34 \pm 0.19 \downarrow$ & $82.23 \pm 0.41 \downarrow$ & $82.63 \pm 0.44$ & $83.42 \pm 0.83$ & $\textbf{84.81} \pm 0.48$ & $82.83 \pm 0.80$ \\
 &  & MF & $78.13 \pm 0.80$ & $78.25 \pm 0.69$ & $77.99 \pm 0.62 \downarrow$ & $74.46 \pm 0.11 \downarrow$ & $76.52 \pm 0.49 \downarrow$ & $78.83 \pm 1.73$ & $78.80 \pm 0.77$ & $\textbf{80.44} \pm 1.09$ & $78.61 \pm 1.05$ \\
 &  & Gap & $6.58 \pm 0.92$ & $6.66 \pm 0.76$ & $6.99 \pm 0.87$ & $7.61 \pm 0.06$ & $8.01 \pm 1.04 $ & $6.28 \pm 2.16$ & $6.97 \pm 0.41$ & $7.17 \pm 1.16$ & $\textbf{6.12} \pm 0.98$ \\

\hline
\end{tabular}
}
\caption{Classification results on medical datasets with standard deviation.}
\label{table:comparison_medical_full}
\end{table*}

\begin{table*}[h!]
\setlength{\tabcolsep}{1mm}
\centering
\renewcommand{\arraystretch}{1.2}
\resizebox{\linewidth}{!}{

\begin{tabular}{lcc|ccccccc|cc}
\hline
\textbf{Dataset} & \textbf{Sensitive} & \textbf{Metric} & \textbf{ERM} & \textbf{Adversarial} & \textbf{CFL} & \textbf{FSCL+} & \textbf{FIN} & \textbf{GroupDRO} & \textbf{FIS} & \textbf{Decoupled} & \textbf{FairSDE} \\
\hline
\multirow{3}{*}{CelebA-Hair} & \multirow{3}{*}{Gender} & ACC & $81.28 \pm 0.05$ & $79.67 \pm 0.58 \downarrow$ & $81.49 \pm 0.08$ & $81.69 \pm 0.33$ & $81.13 \pm 0.02 \downarrow$ & $79.63 \pm 0.25 \downarrow$ & $80.14 \pm 0.06 \downarrow$ & $91.14 \pm 0.06 \downarrow$ & $\textbf{82.54} \pm 0.15$ \\
 &  & MF & $77.60 \pm 0.09$ & $75.71 \pm 1.63 \downarrow$ & $77.72 \pm 0.04$ & $78.35 \pm 0.36$ & $77.28 \pm 0.14 \downarrow$ & $78.35 \pm 0.19$ & $76.64 \pm 0.15 \downarrow$ & $89.76 \pm 0.18 \downarrow$ & $\textbf{79.16} \pm 0.05$ \\
 &  & Gap & $6.32 \pm 0.11$ & $6.80 \pm 1.95$ & $6.48 \pm 0.09$ & $5.73 \pm 0.39$ & $6.60 \pm 0.28$ & $\textbf{2.20} \pm 0.18$ & $6.01 \pm 0.22$ & $2.36 \pm 0.24$ & $5.80 \pm 0.30$ \\
\cline{2-12}
\multirow{3}{*}{CelebA-Smiling} & \multirow{3}{*}{Gender} & ACC & $91.87 \pm 0.04$ & $90.32 \pm 0.95 \downarrow$ & $91.88 \pm 0.15$ & $91.76 \pm 0.10 \downarrow$ & $91.73 \pm 0.09 \downarrow$ & $91.50 \pm 0.19 \downarrow$  & $91.14 \pm 0.06 \downarrow$ & $91.82 \pm 0.04 \downarrow$ & $\textbf{92.22} \pm 0.07$ \\
 &  & MF & $90.68 \pm 0.04$ & $89.19 \pm 0.64 \downarrow$ & $90.64 \pm 0.24 \downarrow$ & $90.72 \pm 0.36$ & $90.53 \pm 0.18 \downarrow$ & $90.98 \pm 0.21$ & $89.76 \pm 0.18 \downarrow$ & $90.58 \pm 0.17 \downarrow$ & $\textbf{91.21} \pm 0.15$ \\
 &  & Gap & $2.04 \pm 0.12$ & $1.92 \pm 1.20$ & $2.12 \pm 0.29 $ & $1.77 \pm 0.52$ & $2.06 \pm 0.25 $ & $\textbf{0.88} \pm 0.06$ & $2.36 \pm 0.24$ & $2.12 \pm 0.17 $ & ${1.71} \pm 0.16$ \\
\hline
\multirow{6}{*}{UTK} & \multirow{3}{*}{Gender} & ACC & $78.06 \pm 0.55$ & $73.40 \pm 3.76 \downarrow$ & $77.89 \pm 0.54 \downarrow$ & $79.97 \pm 0.25$ & $73.93 \pm 1.90 \downarrow$ & $77.84 \pm 0.43 \downarrow$ & $76.32 \pm 1.59 \downarrow$ & $75.50 \pm 0.66 \downarrow$ & $\textbf{80.45} \pm 0.34$ \\
 &  & MF & $76.52 \pm 0.74$ & $71.95 \pm 3.28 \downarrow$ & $76.21 \pm 0.33 \downarrow$ & $78.62 \pm 0.58$ & $70.89 \pm 1.99 \downarrow$ & $76.83 \pm 0.31$ & $74.86 \pm 2.16 \downarrow$ & $72.58 \pm 1.13 \downarrow$ & $\textbf{79.52} \pm 0.70$ \\
 &  & Gap & $3.29 \pm 0.47$ & $3.27 \pm 1.40$ & $3.60 \pm 0.50$ & $2.87 \pm 0.79$ & $6.49 \pm 0.24$ & $2.15 \pm 0.67$ & $3.11 \pm 1.57$ & $6.23 \pm 1.05$ & $\textbf{1.98} \pm 1.41$ \\
\cline{2-12}
 & \multirow{3}{*}{Race} & ACC & $81.65 \pm 0.45$ & $78.39 \pm 3.10 \downarrow$ & $82.84 \pm 0.14$ & ${83.62} \pm 0.41$ & $78.86 \pm 1.38 \downarrow$ & $82.83 \pm 1.21$ & $81.61 \pm 0.53 \downarrow$ & $81.39 \pm 0.58 \downarrow$ & $\textbf{84.17} \pm 0.37$ \\
 &  & MF & $80.04 \pm 0.63$ & $77.10 \pm 3.04 \downarrow$ & $81.29 \pm 0.39$ & ${82.08} \pm 0.26$ & $77.80 \pm 1.52 \downarrow$ & $81.62 \pm 1.48$ & $80.02 \pm 0.51 \downarrow$  & $80.52 \pm 0.52 $ & $\textbf{82.75} \pm 0.19$ \\
 &  & Gap & $3.81 \pm 0.44$ & $3.04 \pm 0.74$ & $3.68 \pm 0.62$ & $3.67 \pm 1.43$ & $2.51 \pm 0.39$ & $2.86 \pm 0.73$ & $3.77 \pm 0.20$ & $\textbf{2.06} \pm 0.56$ & $3.37 \pm 0.73$ \\
\hline
\end{tabular}
}
\caption{Classification results on face datasets with standard deviation.}
\label{table:comparison_face_full}
\end{table*}

\begin{figure}[t]
    \centering
    \includegraphics[width=\textwidth]{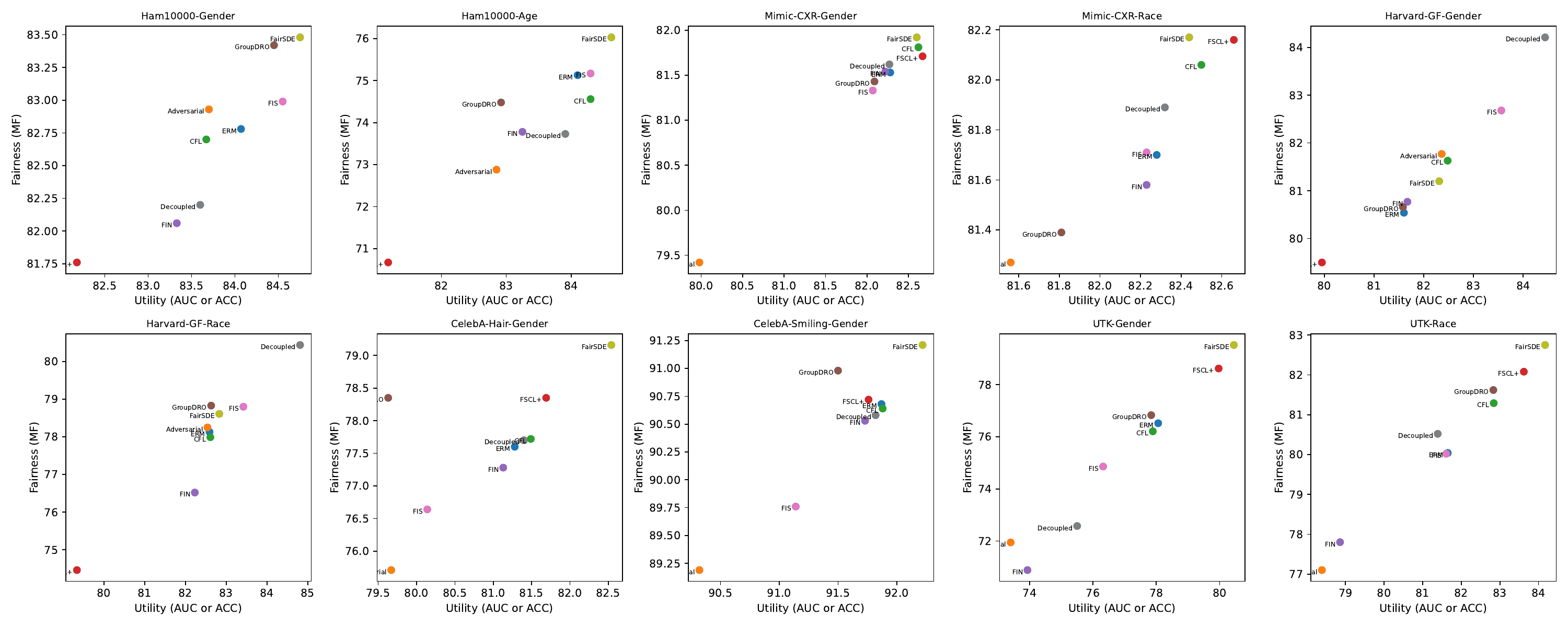}
    \caption{AUC/ACC–Fairness (MF) tradeoff across all datasets.}
    \label{fig:auc_fairness_tradeoff}
\end{figure}

To further illustrate the fairness–performance relationship, we use MF as an example to present the AUC/ACC–Fairness (MF) tradeoff across all datasets and sensitive attributes in Figure~\ref{fig:auc_fairness_tradeoff}. This visualization highlights how different methods balance predictive performance and fairness. A superior algorithm should appear at the outermost region of the curve, pushing the Pareto frontier outward. Our results demonstrate that FairSDE lies closer to the Pareto-optimal boundary in most cases, indicating a better tradeoff between utility and fairness compared with other baselines.

\end{document}